# Cost-Sensitive Classification: Empirical Evaluation of a Hybrid Genetic Decision Tree Induction Algorithm

**Peter D. Turney**                                    TURNEY@AI.IIT.NRC.CA
*Knowledge Systems Laboratory, Institute for Information Technology*
*National Research Council Canada, Ottawa, Ontario, Canada, K1A 0R6.*

## Abstract

This paper introduces ICET, a new algorithm for cost-sensitive classification. ICET uses a genetic algorithm to evolve a population of biases for a decision tree induction algorithm. The fitness function of the genetic algorithm is the average cost of classification when using the decision tree, including both the costs of tests (features, measurements) and the costs of classification errors. ICET is compared here with three other algorithms for cost-sensitive classification — EG2, CS-ID3, and IDX — and also with C4.5, which classifies without regard to cost. The five algorithms are evaluated empirically on five real-world medical datasets. Three sets of experiments are performed. The first set examines the baseline performance of the five algorithms on the five datasets and establishes that ICET performs significantly better than its competitors. The second set tests the robustness of ICET under a variety of conditions and shows that ICET maintains its advantage. The third set looks at ICET's search in bias space and discovers a way to improve the search.

## 1. Introduction

The prototypical example of the problem of cost-sensitive classification is medical diagnosis, where a doctor would like to balance the costs of various possible medical tests with the expected benefits of the tests for the patient. There are several aspects to this problem: When does the benefit of a test, in terms of more accurate diagnosis, justify the cost of the test? When is it time to stop testing and make a commitment to a particular diagnosis? How much time should be spent pondering these issues? Does an extensive examination of the various possible sequences of tests yield a significant improvement over a simpler, heuristic choice of tests? These are some of the questions investigated here.

The words "cost", "expense", and "benefit" are used in this paper in the broadest sense, to include factors such as quality of life, in addition to economic or monetary cost. Cost is domain-specific and is quantified in arbitrary units. It is assumed here that the costs of tests are measured in the same units as the benefits of correct classification. Benefit is treated as negative cost.

This paper introduces a new algorithm for cost-sensitive classification, called ICET (Inexpensive Classification with Expensive Tests — pronounced "iced tea"). ICET uses a genetic algorithm (Grefenstette, 1986) to evolve a population of biases for a decision tree induction algorithm (a modified version of C4.5, Quinlan, 1992). The fitness function of the genetic algorithm is the average cost of classification when using the decision tree, including both the costs of tests (features, measurements) and the costs of classification errors. ICET has the following features: (1) It is sensitive to test costs. (2) It is sensitive to classification error costs. (3) It combines a greedy search heuristic with a genetic search algorithm. (4) It can handle conditional costs, where the cost of one test is conditional on whether a second test has been selected yet. (5) It distinguishes tests with immediate results from tests with delayed results.

The problem of cost-sensitive classification arises frequently. It is a problem in medical diagnosis (Núñez, 1988, 1991), robotics (Tan & Schlimmer, 1989, 1990; Tan, 1993), industrial production processes (Verdenius, 1991), communication network troubleshooting (Lirov & Yue, 1991), machinery diagnosis (where the main cost is skilled labor), automated testing of electronic equipment (where the main cost is time), and many other areas.

There are several machine learning algorithms that consider the costs of tests, such as EG2 (Núñez, 1988, 1991), CS-ID3 (Tan & Schlimmer, 1989, 1990; Tan, 1993), and IDX (Norton, 1989). There are also several algorithms that consider the costs of classification errors (Breiman *et al.*, 1984; Friedman & Stuetzle, 1981; Hermans *et al.*, 1974; Gordon & Perlis, 1989; Pazzani *et al.*, 1994; Provost, 1994; Provost & Buchanan, in press; Knoll *et al.*, 1994). However, there is very little work that considers both costs together.

There are good reasons for considering both the costs of tests and the costs of classification errors. An agent cannot rationally determine whether a test should be performed without knowing the costs of correct and incorrect classification. An agent must balance the cost of each test with the contribution of the test to accurate classification. The agent must also consider when further testing is not economically justified. It often happens that the benefits of further testing are not worth the costs of the tests. This means that a cost must be assigned to both the tests and the classification errors.

Another limitation of many existing cost-sensitive classification algorithms (EG2, CS-ID3) is that they use *greedy* heuristics, which select at each step whatever test contributes most to accuracy and least to cost. A more sophisticated approach would evaluate the interactions among tests in a sequence of tests. A test that appears useful considered in isolation, using a greedy heuristic, may not appear as useful when considered in combination with other tests. Past work has demonstrated that more sophisticated algorithms can have superior performance (Tcheng *et al.*, 1989; Ragavan & Rendell, 1993; Norton, 1989; Schaffer, 1993; Rymon, 1993; Seshu, 1989; Provost, 1994; Provost & Buchanan, in press).

Section 2 discusses why a decision tree is the natural form of knowledge representation for classification with expensive tests and how we measure the average cost of classification of a decision tree. Section 3 introduces the five algorithms that we examine here, C4.5 (Quinlan, 1992), EG2 (Núñez, 1991), CS-ID3 (Tan & Schlimmer, 1989, 1990; Tan, 1993), IDX (Norton, 1989), and ICET. The five algorithms are evaluated empirically on five real-world medical datasets. The datasets are discussed in detail in Appendix A. Section 4 presents three sets of experiments. The first set (Section 4.1) of experiments examines the baseline performance of the five algorithms on the five datasets and establishes that ICET performs significantly better than its competitors for the given datasets. The second set (Section 4.2) tests the robustness of ICET under a variety of conditions and shows that ICET maintains its advantage. The third set (Section 4.3) looks at ICET's search in bias space and discovers a way to improve the search. We then discuss related work and future work in Section 5. We end with a summary of what we have learned with this research and a statement of the general motivation for this type of research.

## 2. Cost-Sensitive Classification

This section first explains why a decision tree is the natural form of knowledge representation for classification with expensive tests. It then discusses how we measure the average cost of classification of a decision tree. Our method for measuring average cost handles aspects of the problem that are typically ignored. The method can be applied to any standard classification decision tree, regardless of how the tree is generated. We end with a discussion of the relation between cost and accuracy.

## 2.1 Decision Trees and Cost-Sensitive Classification

The decision trees used in decision theory (Pearl, 1988) are somewhat different from the classification decision trees that are typically used in machine learning (Quinlan, 1992). When we refer to decision trees in this paper, we mean the standard classification decision trees of machine learning. The claims we make here about classification decision trees also apply to decision theoretical decision trees, with some modification. A full discussion of decision theoretical decision trees is outside of the scope of this paper.

The decision to do a test must be based on both the cost of tests and the cost of classification errors. If a test costs $10 and the maximum penalty for a classification error is $5, then there is clearly no point in doing the test. On the other hand, if the penalty for a classification error is $10,000, the test may be quite worthwhile, even if its information content is relatively low. Past work with algorithms that are sensitive to test costs (Núñez, 1988, 1991; Tan, 1993; Norton, 1989) has overlooked the importance of also considering the cost of classification errors.

When tests are inexpensive, relative to the cost of classification errors, it may be rational to do all tests (i.e., measure all features; determine the values of all attributes) that seem possibly relevant. In this kind of situation, it is convenient to separate the selection of tests from the process of making a classification. First we can decide on the set of tests that are relevant, then we can focus on the problem of learning to classify a case, using the results of these tests. This is a common approach to classification in the machine learning literature. Often a paper focuses on the problem of learning to classify a case, without any mention of the decisions involved in selecting the set of relevant tests.[1]

When tests are expensive, relative to the cost of classification errors, it may be suboptimal to separate the selection of tests from the process of making a classification. We may be able to achieve much lower costs by interleaving the two. First we choose a test, then we examine the test result. The result of the test gives us information, which we can use to influence our choice for the next test. At some point, we decide that the cost of further tests is not justified, so we stop testing and make a classification.

When the selection of tests is interleaved with classification in this way, a decision tree is the natural form of representation. The root of the decision tree represents the first test that we choose. The next level of the decision tree represents the next test that we choose. The decision tree explicitly shows how the outcome of the first test determines the choice of the second test. A leaf represents the point at which we decide to stop testing and make a classification.

Decision theory can be used to define what constitutes an optimal decision tree, given (1) the costs of the tests, (2) the costs of classification errors, (3) the conditional probabilities of test results, given sequences of prior test results, and (4) the conditional probabilities of classes, given sequences of test results. However, searching for an optimal tree is infeasible (Pearl, 1988). ICET was designed to find a good (but not necessarily optimal) tree, where "good" is defined as "better than the competition" (i.e., IDX, CS-ID3, and EG2).

---

1. Not all papers are like this. Decision tree induction algorithms such as C4.5 (Quinlan, 1992) automatically select relevant tests. Aha and Bankert (1994), among others, have used sequential test selection procedures in conjunction with a supervised learning algorithm.

## 2.2 Calculating the Average Cost of Classification

In this section, we describe how we calculate the average cost of classification for a decision tree, given a set of testing data. The method described here is applied uniformly to the decision trees generated by the five algorithms examined here (EG2, CS-ID3, IDX, C4.5, and ICET). The method assumes only a standard classification decision tree (such as generated by C4.5); it makes no assumptions about how the tree is generated. The purpose of the method is to give a plausible estimate of the average cost that can be expected in a real-world application of the decision tree.

We assume that the dataset has been split into a training set and a testing set. The expected cost of classification is estimated by the average cost of classification for the testing set. The average cost of classification is calculated by dividing the total cost for the whole testing set by the number of cases in the testing set. The total cost includes both the costs of tests and the costs of classification errors. In the simplest case, we assume that we can specify test costs simply by listing each test, paired with its corresponding cost. More complex cases will be considered later in this section. We assume that we can specify the costs of classification errors using a classification cost matrix.

Suppose there are $c$ distinct classes. A classification cost matrix is a $c \times c$ matrix, where the element $C_{i,j}$ is the cost of guessing that a case belongs in class $i$, when it actually belongs in class $j$. We do not need to assume any constraints on this matrix, except that costs are finite, real values. We allow negative costs, which can be interpreted as benefits. However, in the experiments reported here, we have restricted our attention to classification cost matrices in which the diagonal elements are zero (we assume that correct classification has no cost) and the off-diagonal elements are positive numbers.[2]

To calculate the cost of a particular case, we follow its path down the decision tree. We add up the cost of each test that is chosen (i.e., each test that occurs in the path from the root to the leaf). If the same test appears twice, we only charge for the first occurrence of the test. For example, one node in a path may say "patient age is less than 10 years" and another node may say "patient age is more than 5 years", but we only charge once for the cost of determining the patient's age. The leaf of the tree specifies the tree's guess for the class of the case. Given the actual class of the case, we use the cost matrix to determine the cost of the tree's guess. This cost is added to the costs of the tests, to determine the total cost of classification for the case.

This is the core of our method for calculating the average cost of classification of a decision tree. There are two additional elements to the method, for handling conditional test costs and delayed test results.

We allow the cost of a test to be conditional on the choice of prior tests. Specifically, we consider the case where a group of tests shares a common cost. For example, a set of blood tests shares the common cost of collecting blood from the patient. This common cost is charged only once, when the decision is made to do the first blood test. There is no charge for collecting blood for the second blood test, since we may use the blood that was collected for the first blood test. Thus the cost of a test in this group is conditional on whether another member of the group has already been chosen.

Common costs appear frequently in testing. For example, in diagnosis of an aircraft engine, a group of tests may share the common cost of removing the engine from the plane and installing it in a test cell. In semiconductor manufacturing, a group of tests may share the common cost of reserving a region on the silicon wafer for a special test structure. In image recognition, a group of image processing algorithms may share a common preprocessing algorithm. These examples show that a realistic assessment of the cost of using a decision tree will frequently need to make allowances for conditional test costs.

---

2. This restriction seems reasonable as a starting point for exploring cost-sensitive classification. In future work, we will investigate the effects of weakening the restriction.

It often happens that the result of a test is not available immediately. For example, a medical doctor typically sends a blood test to a laboratory and gets the result the next day. We allow a test to be labelled either "immediate" or "delayed". If a test is delayed, we cannot use its outcome to influence the choice of the next test. For example, if blood tests are delayed, then we cannot allow the outcome of one blood test to play a role in the decision to do a second blood test. We must make a commitment to doing (or not doing) the second blood test before we know the results of the first blood test.

Delayed tests are relatively common. For example, many medical tests must be shipped to a laboratory for analysis. In gas turbine engine diagnosis, the main fuel control is frequently shipped to a specialized company for diagnosis or repair. In any classification problem that requires multiple experts, one of the experts might not be immediately available.

We handle immediate tests in a decision tree as described above. We handle delayed tests as follows. We follow the path of a case from the root of the decision tree to the appropriate leaf. If we encounter a node, anywhere along this path, that is a delayed test, we are then committed to performing all of the tests in the subtree that is rooted at this node. Since we cannot make the decision to do tests below this node conditional on the outcome of the test at this node, we must pledge to pay for all the tests that we might possibly need to perform, from this point onwards in the decision tree.

Our method for handling delayed tests may seem a bit puzzling at first. The difficulty is that a decision tree combines a method for selecting tests with a method for classifying cases. When tests are delayed, we are forced to proceed in two phases. In the first phase, we select tests. In the second phase, we collect test results and classify the case. For example, a doctor collects blood from a patient and sends the blood to a laboratory. The doctor must tell the laboratory what tests are to be done on the blood. The next day, the doctor gets the results of the tests from the laboratory and then decides on the diagnosis of the patient. A decision tree does not naturally handle a situation like this, where the selection of tests is isolated from the classification of cases. In our method, in the first phase, the doctor uses the decision tree to select the tests. As long as the tests are immediate, there is no problem. As soon as the first delayed test is encountered, the doctor must select all the tests that might possibly be needed in the second phase.[3] That is, the doctor must select all the tests in the subtree rooted at the first delayed test. In the second phase, when the test results arrive the next day, the doctor will have all the information required to go from the root of the tree to a leaf, to make a classification. The doctor must pay for *all* of the tests in the subtree, even though only the tests along one branch of the subtree will actually be used. The doctor does not know in advance *which* branch will actually be used, at the time when it is necessary to order the blood tests. The laboratory that does the blood tests will naturally want the doctor to pay for all the tests that were ordered, even if they are not all used in making the diagnosis.

In general, it makes sense to do all of the desired immediate tests before we do any of the desired delayed tests, since the outcome of an immediate test can be used to influence the decision to do a delayed test, but not vice versa. For example, a medical doctor will question a patient (questions are immediate tests) before deciding what blood tests to order (blood tests are delayed tests).[4]

---

3. This is a simplification of the situation in the real world. A more realistic treatment of delayed tests is one of the areas for future work (Section 5.2).

When all of the tests are delayed (as they are in the BUPA data in Appendix A.1), we must decide in advance (before we see any test results) what tests are to be performed. For a given decision tree, the total cost of tests will be the same for all cases. In situations of this type, the problem of minimizing cost simplifies to the problem of choosing the best subset of the set of available tests (Aha and Bankert, 1994). The sequential order of the tests is no longer important for reducing cost.

Let us consider a simple example to illustrate the method. Table 1 shows the test costs for four tests. Two of the tests are immediate and two are delayed. The two delayed tests share a common cost of $2.00. There are two classes, 0 and 1. Table 2 shows the classification cost matrix. Figure 1 shows a decision tree. Table 3 traces the path through the tree for a particular case and shows how the cost is calculated. The first step is to do the test at the root of the tree (test alpha). In the second step, we encounter a delayed test (delta), so we must calculate the cost of the entire subtree rooted at this node. Note that epsilon only costs $8.00, since we have already selected delta, and delta and epsilon have a common cost. In the third step, we do test epsilon, but we do not need to pay, since we already paid in the second step. In the fourth step, we guess the class of the case. Unfortunately, we guess incorrectly, so we pay a penalty of $50.00.

Table 1: Test costs for a simple example.

|   | Test | Group | Cost | Delayed |
|---|------|-------|------|---------|
| 1 | alpha |  | $5.00 | no |
| 2 | beta |  | $10.00 | no |
| 3 | delta | A | $7.00 if first test in group A, $5.00 otherwise | yes |
| 4 | epsilon | A | $10.00 if first test in group A, $8.00 otherwise | yes |

Table 2: Classification costs for a simple example.

| Actual Class | Guess Class | Cost |
|--------------|-------------|------|
| 0 | 0 | $0.00 |
| 0 | 1 | $50.00 |
| 1 | 0 | $50.00 |
| 1 | 1 | $0.00 |

---

4. In the real world, there are many factors that can influence the sequence of tests, such as the length of the delay and the probability that the delayed test will be needed. When we ignore these many factors and pay attention only to the simplified model presented here, it makes sense to do all of the desired immediate tests before we do any of the desired delayed tests. We do not know to what extent this actually occurs in the real world. One complication is that medical doctors in most industrialized countries are not directly affected by the cost of the tests they select. In fact, fear of law suits gives them incentive to order unnecessary tests.

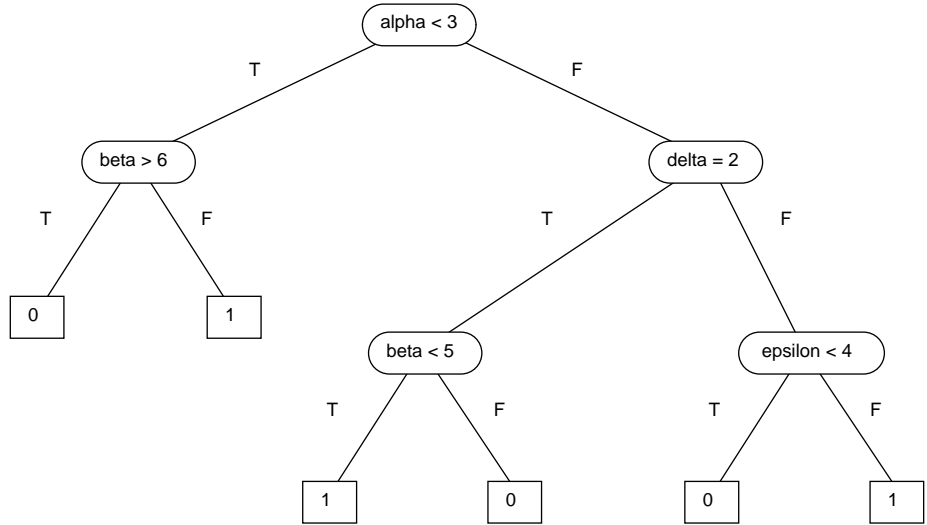

Figure 1: Decision tree for a simple example.

Table 3: Calculating the cost for a particular case.

| Step | Action | Result | Cost |
|------|--------|--------|------|
| 1 | do alpha | alpha = 6 | $5.00 |
| 2 | do delta | delta = 3 | $7.00 + $10.00 + $8.00 = $25.00 |
| 3 | do epsilon | epsilon = 2 | already paid, in step #2 |
| 4 | guess class = 0 | actual class = 1 | $50.00 |
| total cost | | | $80.00 |

In summary, this section presents a method for estimating the average cost of using a given decision tree. The decision tree can be any standard classification decision tree; no special assumptions are made about the tree; it does not matter how the tree was generated. The method requires (1) a decision tree (Figure 1), (2) information on the calculation of test costs (Table 1), (3) a classification cost matrix (Table 2), and (4) a set of testing data (Table 3). The method is (i) sensitive to the cost of tests, (ii) sensitive to the cost of classification errors, (iii) capable of handling conditional test costs, and (iv) capable of handling delayed tests. In the experiments reported in Section 4, this method is applied uniformly to all five algorithms.

## 2.3 Cost and Accuracy

Our method for calculating cost does not explicitly deal with accuracy; however, we can handle accuracy as a special case. If the test cost is set to $0.00 for all tests and the classification cost matrix is set to a positive constant value $k$ when the guess class $i$ does not equal the actual class $j$, but it is set to $0.00 when $i$ equals $j$, then the average total cost of using the decision tree is $pk$, where $p \in [0,1]$ is the frequency of errors on the testing dataset and $100\,(1-p)$ is the percentage accuracy on the testing dataset. Thus there is a linear relationship between average total cost and percentage accuracy, in this situation.

More generally, let $C$ be a classification cost matrix that has cost $x$ on the diagonal, $C_{i,i} = x$, and cost $y$ off the diagonal, $(i \neq j) \rightarrow (C_{i,j} = y)$, where $x$ is less than $y$, $x < y$. We will call this type of classification cost matrix a *simple classification cost matrix*. A cost matrix that is not simple will be called a *complex classification cost matrix*.[5] When we have a simple cost matrix and test costs are zero (equivalently, test costs are ignored), minimizing cost is exactly equivalent to maximizing accuracy.

It follows from this that an algorithm that is sensitive to misclassification error costs but ignores test costs (Breiman *et al.*, 1984; Friedman & Stuetzle, 1981; Hermans *et al.*, 1974; Gordon & Perlis, 1989; Pazzani *et al.*, 1994; Provost, 1994; Provost & Buchanan, in press; Knoll *et al.*, 1994) will only be interesting when we have a complex cost matrix. If we have a simple cost matrix, an algorithm such as CART (Breiman *et al.*, 1984) that is sensitive to misclassification error cost has no advantage over an algorithm such as C4.5 (Quinlan, 1992) that maximizes accuracy (assuming other differences between these two algorithms are negligible). Most of the experiments in this paper use a simple cost matrix (the only exception is Section 4.2.3). Therefore we focus on comparison of ICET with algorithms that are sensitive to test cost (IDX, CS-ID3, and EG2). In future work, we will examine complex cost matrices and compare ICET with algorithms that are sensitive to misclassification error cost.

It is difficult to find information on the costs of misclassification errors in medical practice, but it seems likely that a complex cost matrix is more appropriate than a simple cost matrix for most medical applications. This paper focuses on simple cost matrices because, as a research strategy, it seems wise to start with the simple cases before we attempt the complex cases.

Provost (Provost, 1994; Provost & Buchanan, in press) combines accuracy and classification error cost using the following formula:

$$\text{score} = A \cdot \text{accuracy} - B \cdot \text{cost} \tag{1}$$

In this formula, $A$ and $B$ are arbitrary weights that the user can set for a particular application. Both "accuracy" and "cost", as defined by Provost (Provost, 1994; Provost & Buchanan, in press), can be represented using classification cost matrices. We can represent "accuracy" using any simple cost matrix. In interesting applications, "cost" will be represented by a complex cost matrix. Thus "score" is a weighted sum of two classification cost matrices, which means that "score" is itself a classification cost matrix. This shows that equation (1) can be handled as a special case of the method presented here. There is no loss of information in this translation of Provost's formula into a cost matrix. This does not mean that *all* criteria can be represented as costs. An example of a criterion that cannot be represented as a cost is *stability* (Turney, in press).

## 3. Algorithms

This section discusses the algorithms used in this paper: C4.5 (Quinlan, 1992), EG2 (Núñez, 1991), CS-ID3 (Tan & Schlimmer, 1989, 1990; Tan, 1993), IDX (Norton, 1989), and ICET.

---

5. We will occasionally say "simple cost matrix" or "complex cost matrix". This should not cause confusion, since test costs are not represented with a matrix.

### 3.1 C4.5

C4.5 (Quinlan, 1992) builds a decision tree using the standard TDIDT (top-down induction of decision trees) approach, recursively partitioning the data into smaller subsets, based on the value of an attribute. At each step in the construction of the decision tree, C4.5 selects the attribute that maximizes the information gain ratio. The induced decision tree is pruned using pessimistic error estimation (Quinlan, 1992). There are several parameters that can be adjusted to alter the behavior of C4.5. In our experiments with C4.5, we used the default settings for all parameters. We used the C4.5 source code that is distributed with (Quinlan, 1992).

### 3.2 EG2

EG2 (Núñez, 1991) is a TDIDT algorithm that uses the Information Cost Function (ICF) (Núñez, 1991) for selection of attributes. ICF selects attributes based on both their information gain and their cost. We implemented EG2 by modifying the C4.5 source code so that ICF was used instead of information gain ratio.

ICF for the $i$-th attribute, $\text{ICF}_i$, is defined as follows:[6]

$$\text{ICF}_i = \frac{2^{\Delta I_i} - 1}{(C_i + 1)^\omega} \qquad \text{where } 0 \le \omega \le 1 \tag{2}$$

In this equation, $\Delta I_i$ is the information gain associated with the $i$-th attribute at a given stage in the construction of the decision tree and $C_i$ is the cost of measuring the $i$-th attribute. C4.5 selects the attribute that maximizes the information gain ratio, which is a function of the information gain $\Delta I_i$. We modified C4.5 so that it selects the attribute that maximizes $\text{ICF}_i$.

The parameter $\omega$ adjusts the strength of the bias towards lower cost attributes. When $\omega = 0$, cost is ignored and selection by $\text{ICF}_i$ is equivalent to selection by $\Delta I_i$. When $\omega = 1$, $\text{ICF}_i$ is strongly biased by cost. Ideally, $\omega$ would be selected in a way that is sensitive to classification error cost (this is done in ICET — see Section 3.5). Núñez (1991) does not suggest a principled way of setting $\omega$. In our experiments with EG2, $\omega$ was set to 1. In other words, we used the following selection measure:

$$\frac{2^{\Delta I_i} - 1}{C_i + 1} \tag{3}$$

In addition to its sensitivity to the cost of tests, EG2 generalizes attributes by using an ISA tree (a generalization hierarchy). We did not implement this aspect of EG2, since it was not relevant for the experiments reported here.

### 3.3 CS-ID3

CS-ID3 (Tan & Schlimmer, 1989, 1990; Tan, 1993) is a TDIDT algorithm that selects the attribute that maximizes the following heuristic function:

$$\frac{(\Delta I_i)^2}{C_i} \tag{4}$$

---

6. This is the inverse of ICF, as defined by Núñez (1991). Núñez *minimizes* his criterion. To facilitate comparison with the other algorithms, we use equation (2). This criterion is intended to be *maximized*.

We implemented CS-ID3 by modifying C4.5 so that it selects the attribute that maximizes (4).

CS-ID3 uses a lazy evaluation strategy. It only constructs the part of the decision tree that classifies the current case. We did not implement this aspect of CS-ID3, since it was not relevant for the experiments reported here.

### 3.4 IDX

IDX (Norton, 1989) is a TDIDT algorithm that selects the attribute that maximizes the following heuristic function:

$$\frac{\Delta I_i}{C_i} \qquad (5)$$

We implemented IDX by modifying C4.5 so that it selects the attribute that maximizes (5).

C4.5 uses a *greedy* search strategy that chooses at each step the attribute with the highest information gain ratio. IDX uses a lookahead strategy that looks $n$ tests ahead, where $n$ is a parameter that may be set by the user. We did not implement this aspect of IDX. The lookahead strategy would perhaps make IDX more competitive with ICET, but it would also complicate comparison of the heuristic function (5) with the heuristics (3) and (4) used by EG2 and CS-ID3.

### 3.5 ICET

ICET is a hybrid of a genetic algorithm and a decision tree induction algorithm. The genetic algorithm evolves a population of biases for the decision tree induction algorithm. The genetic algorithm we use is GENESIS (Grefenstette, 1986).[7] The decision tree induction algorithm is C4.5 (Quinlan, 1992), modified to use ICF. That is, the decision tree induction algorithm is EG2, implemented as described in Section 3.2.

ICET uses a two-tiered search strategy. On the bottom tier, EG2 performs a greedy search through the space of decision trees, using the standard TDIDT strategy. On the top tier, GENESIS performs a genetic search through a space of biases. The biases are used to modify the behavior of EG2. In other words, GENESIS controls EG2's preference for one type of decision tree over another.

ICET does not use EG2 the way it was designed to be used. The $n$ costs, $C_i$, used in EG2's attribute selection function, are treated by ICET as bias parameters, not as costs. That is, ICET manipulates the bias of EG2 by adjusting the parameters, $C_i$. In ICET, the values of the bias parameters, $C_i$, have no direct connection to the actual costs of the tests.

Genetic algorithms are inspired by biological evolution. The individuals that are evolved by GENESIS are strings of bits. GENESIS begins with a population of randomly generated individuals (bit strings) and then it measures the "fitness" of each individual. In ICET, an individual (a bit string) represents a bias for EG2. An individual is evaluated by running EG2 on the data, using the bias of the given individual. The "fitness" of the individual is the average cost of classification of the decision tree that is generated by EG2. In the next generation, the population is replaced with new individuals. The new individuals are generated from the previous generation, using mutation and crossover (sex). The fittest individuals in the first generation have the most offspring in the second generation. After a fixed number of generations, ICET halts and its output is the decision tree determined by the fittest individual. Figure 2 gives a sketch of the ICET algorithm.

---

7. We used GENESIS Version 5.0, which is available at URL ftp://ftp.aic.nrl.navy.mil/pub/galist/src/ga/genesis.tar.Z or ftp://alife.santafe.edu/pub/USER-AREA/EC/GA/src/gensis-5.0.tar.gz.

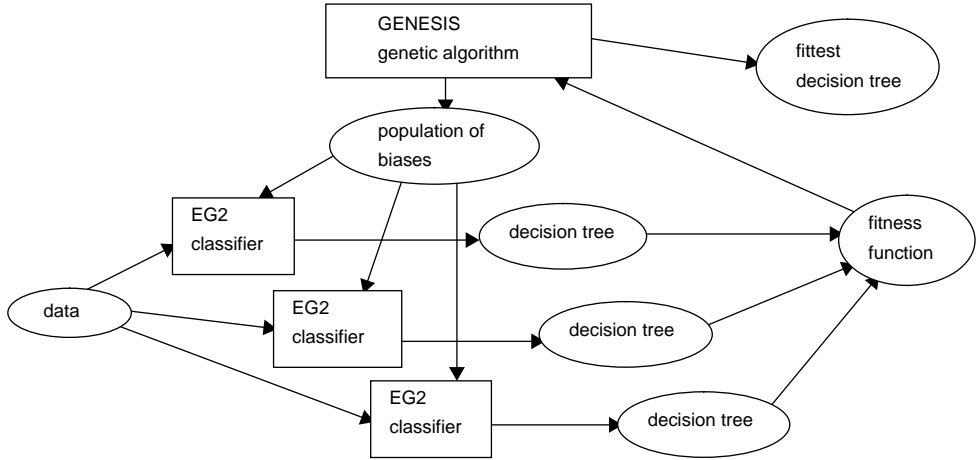

Figure 2: A sketch of the ICET algorithm.

GENESIS has several parameters that can be used to alter its performance. The parameters we used are listed in Table 4. These are essentially the default parameter settings (Grefenstette, 1986). We used a population size of 50 individuals and 1,000 trials, which results in 20 generations. An individual in the population consists of a string of $n + 2$ numbers, where $n$ is the number of attributes (tests) in the given dataset. The $n + 2$ numbers are represented in binary format, using a Gray code.[8] This binary string is used as a bias for EG2. The first $n$ numbers in the string are treated as if they were the $n$ costs, $C_i$, used in ICF (equation (2)). The first $n$ numbers range from 1 to 10,000 and are coded with 12 binary digits each. The last two numbers in the string are used to set ω and CF. The parameter ω is used in ICF. The parameter CF is used in C4.5 to control the level of pruning of the decision tree. The last two numbers are coded with 8 binary digits each. ω ranges from 0 (cost is ignored) to 1 (maximum sensitivity to cost) and CF ranges from 1 (high pruning) to 100 (low pruning). Thus an individual is a string of $12n + 16$ bits.

Each trial of an individual consists of running EG2 (implemented as a modification to C4.5) on a given training dataset, using the numbers specified in the binary string to set $C_i$ ($i = 1, ..., n$), ω, and CF. The training dataset is randomly split into two equal-sized subsets (±1 for odd-sized training sets), a sub-training set and a sub-testing set. A different random split is used for each trial, so the outcome of a trial is stochastic. We cannot assume that identical individuals yield identical outcomes, so every individual must be evaluated. This means that there will be duplicate individuals in the population, with slightly different fitness scores. The measure of fitness of an individual is the average cost of classification on the sub-testing set, using the decision tree that was generated on the sub-training set. The aver-age cost is measured as described in Section 2.2. After 1,000 trials, the most fit (lowest cost) individual is then used as a bias for EG2 with the whole training set as input. The resulting decision tree is the output of ICET for the given training dataset.[9]

---

8. A Gray code is a binary code that is designed to avoid "Hamming cliffs". In the standard binary code, 7 is represented as 0111 and 8 is represented as 1000. These numbers are adjacent, yet the Hamming distance from 0111 to 1000 is large. In a Gray code, adjacent numbers are represented with binary codes that have small Hamming distances. This tends to improve the performance of a genetic algorithm (Grefenstette, 1986).

Table 4: Parameter settings for GENESIS.

| Parameter | Setting |
|-----------|---------|
| Experiments | 1 |
| Total Trials | 1000 |
| Population Size | 50 |
| Structure Length | $12n + 16$ |
| Crossover Rate | 0.6 |
| Mutation Rate | 0.001 |
| Generation Gap | 1.0 |
| Scaling Window | 5 |
| Report Interval | 100 |
| Structures Saved | 1 |
| Max Gens w/o Eval | 2 |
| Dump Interval | 0 |
| Dumps Saved | 0 |
| Options | acefgl |
| Random Seed | 123456789 |
| Rank Min | 0.75 |

The $n$ costs (bias parameters), $C_i$, used in ICF, are not directly related to the true costs of the attributes. The 50 individuals in the first generation are generated randomly, so the initial values of $C_i$ have no relation to the true costs. After 20 generations, the values of $C_i$ may have some relation to the true costs, but it will not be a simple relationship. These values of $C_i$ are more appropriately thought of as *biases* than costs. Thus GENESIS is searching through a bias space for biases for C4.5 that result in decision trees with low average cost.

The biases $C_i$ range from 1 to 10,000. When a bias $C_i$ is greater than 9,000, the $i$-th attribute is ignored. That is, the $i$-th attribute is not available for C4.5 to include in the decision tree, even if it might maximize $ICF_i$. This threshold of 9,000 was arbitrarily chosen. There was no attempt to optimize this value by experimentation.

We chose to use EG2 in ICET, rather than IDX or CS-ID3, because EG2 has the parameter $\omega$, which gives GENESIS greater control over the bias of EG2. $ICF_i$ is partly based on the data (via the information gain, $\Delta I_i$) and it is partly based on the bias (via the "pseudo-cost", $C_i$). The exact mix of data and bias can be controlled by varying $\omega$. Otherwise, there is no reason to prefer EG2 to IDX or CS-ID3, which could easily be used instead of EG2.

---

9. The 50/50 partition of sub-training and sub-testing sets could mean that ICET may not work well on small datasets. The smallest dataset of the five we examine here is the Hepatitis dataset, which has 155 cases. The training sets had 103 cases and the testing sets had 52 cases. The sub-training and sub-testing sets had 51 or 52 cases. We can see from Figure 3 that ICET performed slightly better than the other algorithms on this dataset (the difference is not significant).

The treatment of delayed tests and conditional test costs is not "hard-wired" into EG2. It is built into the fitness function used by GENESIS, the average cost of classification (measured as described in Section 2). This makes it relatively simple to extend ICET to handle other pragmatic constraints on the decision trees.

In effect, GENESIS "lies" to EG2 about the costs of the tests. How can lies improve the performance of EG2? EG2 is a hill-climbing algorithm that can get trapped at a local optimum. It is a greedy algorithm that looks only one test ahead as it builds a decision tree. Because it looks only one step ahead, EG2 suffers from the *horizon effect*. This term is taken from the literature on chess playing programs. Suppose that a chess playing program has a fixed three-move lookahead depth and it finds that it will loose its queen in three moves, if it follows a certain branch of the game tree. There may be an alternate branch where the program first sacrifices a pawn and then loses its queen in four moves. Because the loss of the queen is over its three-move *horizon*, the program may foolishly decide to sacrifice its pawn. One move later, it is again faced with the loss of its queen. Analogously, EG2 may try to avoid a certain expensive test by selecting a less expensive test. One test later, it is again faced with the more expensive test. After it has exhausted all the cheaper tests, it may be forced to do the expensive test, in spite of its efforts to avoid the test. GENESIS can prevent this short-sighted behavior by telling lies to EG2. GENESIS can exaggerate the cost of the cheap tests or it can understate the cost of the expensive test. Based on past trials, GENESIS can find the lies that yield the best performance from EG2.

In ICET, learning (local search in EG2) and evolution (in GENESIS) interact. A common form of hybrid genetic algorithm uses local search to improve the individuals in a population (Schaffer *et al.*, 1992). The improvements are then coded into the strings that represent the individuals. This is a form of *Lamarckian evolution*. In ICET, the improvements due to EG2 are *not* coded into the strings. However, the improvements can accelerate evolution by altering the fitness landscape. This phenomenon (and other phenomena that result from this form of hybrid) is known as the *Baldwin effect* (Baldwin, 1896; Morgan, 1896; Waddington, 1942; Maynard Smith, 1987; Hinton & Nowlan, 1987; Ackley & Littman, 1991; Whitley & Gruau, 1993; Whitley *et al.*, 1994; Anderson, in press). The Baldwin effect may explain much of the success of ICET.

## 4. Experiments

This section describes experiments that were performed on five datasets, taken from the Irvine collection (Murphy & Aha, 1994). The five datasets are described in detail in Appendix A. All five datasets involve medical problems. The test costs are based on information from the Ontario Ministry of Health (1992). The main purpose of the experiments is to gain insight into the behavior of ICET. The other cost-sensitive algorithms, EG2, CS-ID3, and IDX, are included mainly as benchmarks for evaluating ICET. C4.5 is also included as a benchmark, to illustrate the behavior of an algorithm that makes no use of cost information. The main conclusion of these experiments is that ICET performs significantly better than its competitors, under a wide range of conditions. With access to the Irvine collection and the information in Appendix A, it should be possible for other researchers to duplicate the results reported here.

Medical datasets frequently have missing values.[10] We conjecture that many missing values in medical datasets are missing because the doctor involved in generating the dataset decided that a particular test was not economically justified for a particular patient. Thus there may be information content in the fact that a certain value is missing. There may be many reasons for missing values other than the cost of the tests. For example, perhaps the doctor forgot to order the test or perhaps the patient failed to show up for the test. However, it seems likely that there is often information content in the fact that a value is missing. For our experiments, this information content should be hidden from the learning algorithms, since using it (at least in the testing sets) would be a form of cheating. Two of the five datasets we selected had some missing data. To avoid accusations of cheating, we decided to preprocess the datasets so that the data presented to the algorithms had no missing values. This preprocessing is described in Appendices A.2 and A.3.

Note that ICET is capable of handling missing values without preprocessing — it inherits this ability from its C4.5 component. We preprocessed the data only to avoid accusations of cheating, not because ICET requires preprocessed data.

For the experiments, each dataset was randomly split into 10 pairs of training and testing sets. Each training set consisted of two thirds of the dataset and each testing set consisted of the remaining one third. The same 10 pairs were used in all experiments, in order to facilitate comparison of results across experiments.

There are three groups of experiments. The first group of experiments examines the baseline performance of the algorithms. The second group considers how robust ICET is under a variety of conditions. The final group looks at how ICET searches bias space.

### 4.1 Baseline Performance

This section examines the baseline performance of the algorithms. In Section 4.1.1, we look at the average cost of classification of the five algorithms on the five datasets. Averaged across the five datasets, ICET has the lowest average cost. In Section 4.1.2, we study test expenditures and error rates as functions of the penalty for misclassification errors. Of the five algorithms studied here, only ICET adjusts its test expenditures and error rates as functions of the penalty for misclassification errors. The other four algorithms ignore the penalty for misclassification errors. ICET behaves as one would expect, increasing test expenditures and decreasing error rates as the penalty for misclassification errors rises. In Section 4.1.3, we examine the execution time of the algorithms. ICET requires 23 minutes on average on a single-processor Sparc 10. Since ICET is inherently parallel, there is significant room for speed increase on a parallel machine.

#### 4.1.1 Average Cost of Classification

The experiment presented here establishes the baseline performance of the five algorithms. The hypothesis was that ICET will, on average, perform better than the other four algorithms. The classification cost matrix was set to a positive constant value $k$ when the guess class $i$ does not equal the actual class $j$, but it was set to $0.00 when $i$ equals $j$. We experimented with seven settings for $k$, $10, $50, $100, $500, $1000, $5000, and $10000.

Initially, we used the average cost of classification as the performance measure, but we found that there are three problems with using the average cost of classification to compare the five algorithms. First, the differences in costs among the algorithms become relatively small as the penalty for classification errors increases. This makes it difficult to see which algorithm is best. Second, it is difficult to combine the results for the five datasets in a fair manner.[11] It is not fair to average the five datasets together, since their test costs have different scales (see Appendix A). The test costs in the Heart Disease dataset, for example, are substantially larger than the test costs in the other four datasets. Third, it is difficult to combine average costs for different values of $k$ in a fair manner, since more weight will be given to the situations where $k$ is large than to the situations where it is small.

---

10. A survey of 54 datasets from the Irvine collection (URL ftp://ftp.ics.uci.edu/pub/machine-learning-databases/SUMMARY-TABLE) indicates that 85% of the medical datasets (17 out of 20) have missing values, while only 24% (8 out of 34) of the non-medical datasets have missing values.

To address these concerns, we decided to normalize the average cost of classification. We normalized the average cost by dividing it by the *standard cost*. Let $f_i \in [0,1]$ be the frequency of class $i$ in the given dataset. That is, $f_i$ is the fraction of the cases in the dataset that belong in class $i$. We calculate $f_i$ using the entire dataset, not just the training set. Let $C_{i,j}$ be the cost of guessing that a case belongs in class $i$, when it actually belongs in class $j$. Let $T$ be the total cost of doing all of the possible tests. The *standard cost* is defined as follows:

$$T + \min_i (1 - f_i) \cdot \max_{i,j} C_{i,j} \qquad (6)$$

We can decompose formula (6) into three components:

$$T \qquad (7)$$

$$\min_i (1 - f_i) \qquad (8)$$

$$\max_{i,j} C_{i,j} \qquad (9)$$

We may think of (7) as an upper bound on test expenditures, (8) as an upper bound on error rate, and (9) as an upper bound on the penalty for errors. The standard cost is always less than the maximum possible cost, which is given by the following formula:

$$T + \max_{i,j} C_{i,j} \qquad (10)$$

The point is that (8) is not really an upper bound on error rate, since it is possible to be wrong with every guess. However, our experiments suggest that the standard cost is better for normalization, since it is a more realistic (tighter) upper bound on the average cost. In our experiments, the average cost never went above the standard cost, although it occasionally came very close.

Figure 3 shows the result of using formula (6) to normalize the average cost of classification. In the plots, the $x$ axis is the value of $k$ and the $y$ axis is the average cost of classification as a percentage of the standard cost of classification. We see that, on average (the sixth plot in Figure 3), ICET has the lowest classification cost. The one dataset where ICET does not perform particularly well is the Heart Disease dataset (we discuss this later, in Sections 4.3.2 and 4.3.3).

To come up with a single number that characterizes the performance of each algorithm, we averaged the numbers in the sixth plot in Figure 3.[12] We calculated 95% confidence regions for the averages, using the standard deviations across the 10 random splits of the datasets. The result is shown in Table 5.

---

11. We want to combine the results in order to summarize the performance of the algorithms on the five datasets. This is analogous to comparing students by calculating the GPA (Grade Point Average), where students are to courses as algorithms are to datasets.

12. Like the GPA, all datasets (courses) have the same weight. However, unlike the GPA, all algorithms (students) are applied to the same datasets (have taken the same courses). Thus our approach is perhaps more fair to the algorithms than GPA is to students.

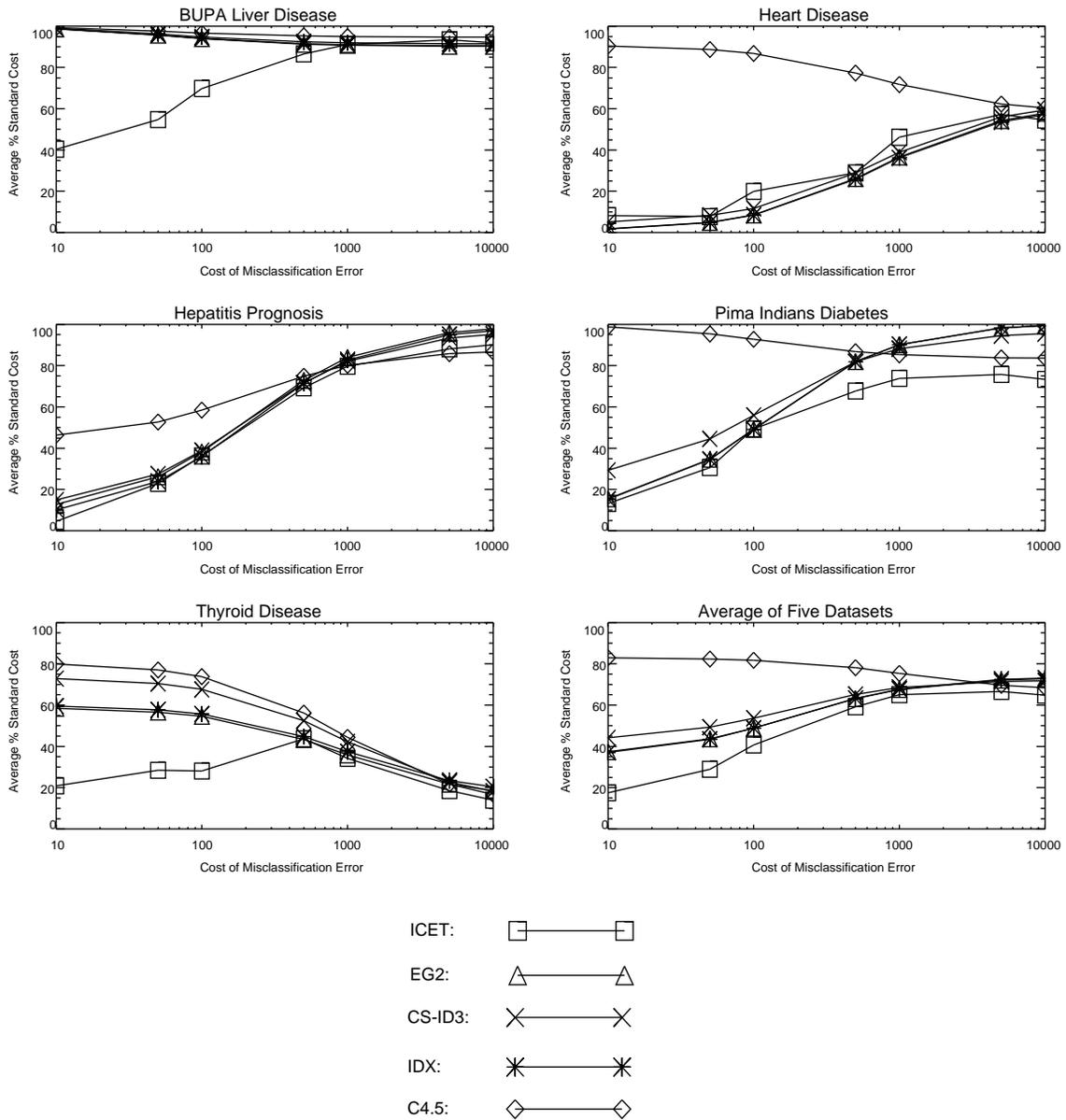

Figure 3: Average cost of classification as a percentage of the standard cost of classification for the baseline experiment.

Table 5 shows the averages for the first three misclassification error costs alone ($10, $50, and $100), in addition to showing the averages for all seven misclassification error costs ($10 to $10000). We have two averages (the two columns in Table 5), based on two groups of data, to address the following argument: As the penalty for misclassification errors increases, the cost of the tests becomes relatively insignificant. With very high misclassification error cost, the test cost is effectively zero, so the task becomes simply to maximize accuracy. As

Table 5: Average percentage of standard cost for the baseline experiment.

| Algorithm | Average Classification Cost as Percentage of Standard ± 95% Confidence | |
|---|---|---|
| | Misclassification Error Costs from 10.00 to 10,000.00 | Misclassification Error Costs from 10.00 to 100.00 |
| ICET | 49 ± 7 | 29 ± 7 |
| EG2 | 58 ± 5 | 43 ± 3 |
| CS-ID3 | 61 ± 6 | 49 ± 4 |
| IDX | 58 ± 5 | 43 ± 3 |
| C4.5 | 77 ± 5 | 82 ± 4 |

we see in Figure 3, the gap between C4.5 (which maximizes accuracy) and the other algorithms becomes smaller as the cost of misclassification error increases. Therefore the benefit of sensitivity to test cost decreases as the cost of misclassification error increases. It could be argued that one would only bother with an algorithm that is sensitive to test cost when tests are relatively expensive, compared to the cost of misclassification errors. Thus the most realistic measure of performance is to examine the average cost of classification when the cost of tests is the same order of magnitude as the cost of misclassification errors ($10 to $100). This is why Table 5 shows both averages.

Our conclusion, based on Table 5, is that ICET performs significantly better than the other four algorithms when the cost of tests is the same order of magnitude as the cost of misclassification errors ($10, $50, and $100). When the cost of misclassification errors dominates the test costs, ICET still performs better than the competition, but the difference is less significant. The other three cost-sensitive algorithms (EG2, CS-ID3, and IDX) perform significantly better than C4.5 (which ignores cost). The performance of EG2 and IDX is indistinguishable, but CS-ID3 appears to be consistently more costly than EG2 and IDX.

### 4.1.2 Test Expenditures and Error Rates as Functions of the Penalty for Errors

We argued in Section 2 that expenditures on tests should be conditional on the penalty for misclassification errors. Therefore ICET is designed to be sensitive to both the cost of tests and the cost of classification errors. This leads us to the hypothesis that ICET tends to spend more on tests as the penalty for misclassification errors increases. We also expect that the error rate of ICET should decrease as test expenditures increase. These two hypotheses are confirmed in Figure 4. In the plots, the $x$ axis is the value of $k$ and the $y$ axis is (1) the average expenditure on tests, expressed as a percentage of the maximum possible expenditure on tests, $T$, and (2) the average percent error rate. On average (the sixth plot in Figure 4), test expenditures rise and error rate falls as the penalty for classification errors increases. There are some minor deviations from this trend, since ICET can only guess at the value of a test (in terms of reduced error rate), based on what it sees in the training dataset. The testing dataset may not always support that guess. Note that plots for the other four algorithms, corresponding to the plots for ICET in Figure 4, would be straight horizontal lines, since all four algorithms ignore the cost of misclassification error. They generate the same decision trees for every possible misclassification error cost.

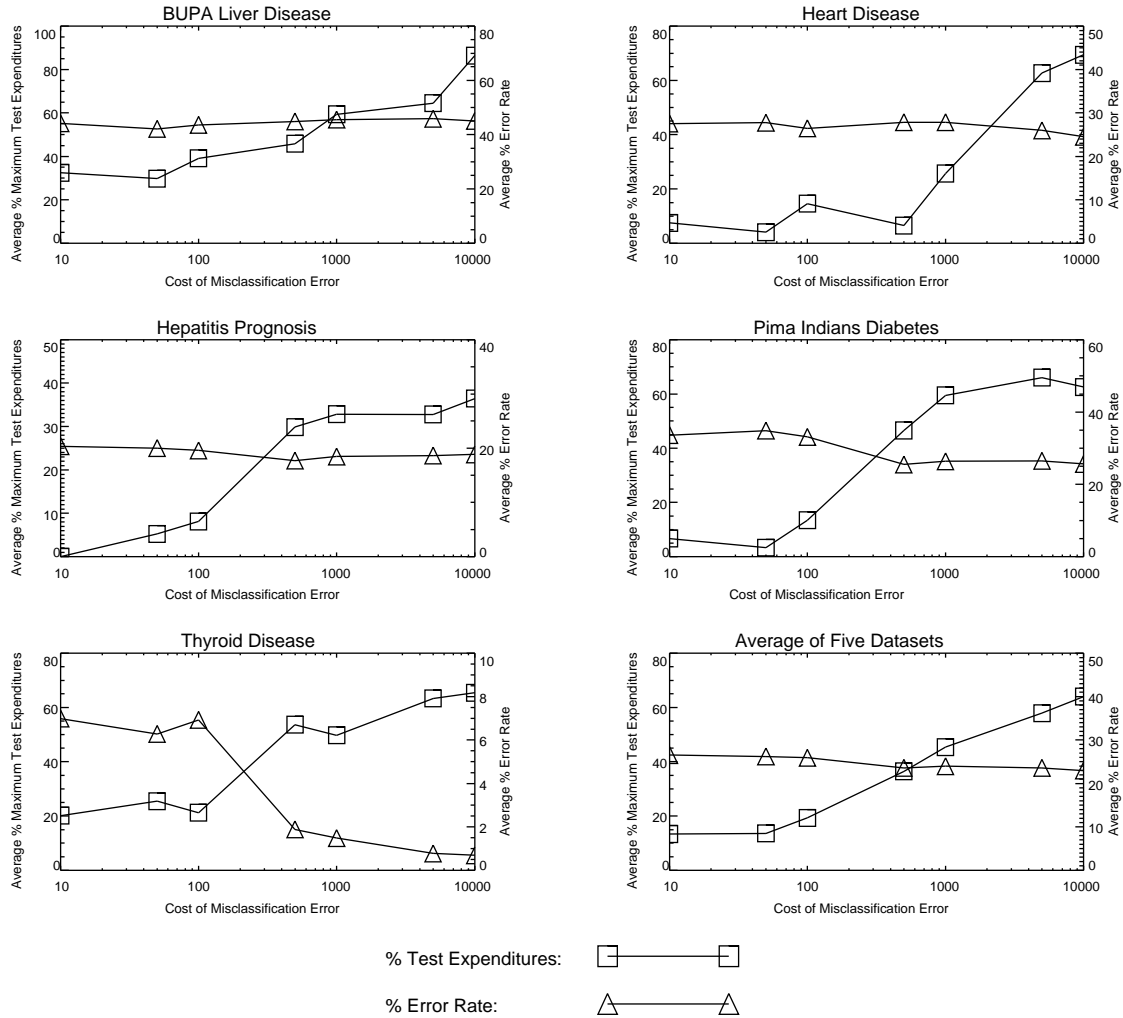

Figure 4: Average test expenditures and average error rate
as a function of misclassification error cost.

### 4.1.3 EXECUTION TIME

In essence, ICET works by invoking C4.5 1000 times (Section 3.5). Fortunately, Quinlan's (1992) implementation of C4.5 is quite fast. Table 6 shows the run-times for the algorithms, using a single-processor Sun Sparc 10. One full experiment takes about one week (roughly 23 minutes for an average run, multiplied by 5 datasets, multiplied by 10 random splits, multiplied by 7 misclassification error costs equals about one week). Since genetic algorithms can easily be executed in parallel, there is substantial room for speed increase with a parallel machine. Each generation consists of 50 individuals, which could be evaluated in parallel, reducing the average run-time to about half a minute.

### 4.2 Robustness of ICET

This group of experiments considers how robust ICET is under a variety of conditions. Each section considers a different variation on the operating environment of ICET. The ICET algorithm itself is not modified. In Section 4.2.1, we alter the environment by labelling all tests as immediate. In Section 4.2.2, we do not recognize shared costs, so there is no discount for a group of tests with a common cost. In Section 4.2.3, we experiment with complex classification cost matrices, where different types of errors have different costs. In Section 4.2.4, we examine what happens when ICET is trained with a certain penalty for misclassification errors, then tested with a different penalty. In all four experiments, we find that ICET continues to perform well.

Table 6: Elapsed run-time for the five algorithms.

| Algorithm | Average Elapsed Run-Time for Each Dataset — Minutes:Seconds | | | | | |
|---|---|---|---|---|---|---|
| | BUPA | Heart | Hepatitis | Pima | Thyroid | Average |
| ICET | 15:43 | 13:14 | 10:29 | 28:19 | 45:25 | 22:38 |
| EG2 | 0:1 | 0:1 | 0:1 | 0:3 | 0:3 | 0:2 |
| CS-ID3 | 0:1 | 0:1 | 0:1 | 0:3 | 0:3 | 0:2 |
| IDX | 0:1 | 0:1 | 0:1 | 0:3 | 0:3 | 0:2 |
| C4.5 | 0:2 | 0:1 | 0:1 | 0:4 | 0:3 | 0:2 |

### 4.2.1 All Tests Immediate

A critic might object that the previous experiments do not show that ICET is superior to the other algorithms due to its sensitivity to both test costs and classification error costs. Perhaps ICET is superior simply because it can handle delayed tests, while the other algorithms treat all tests as immediate.[13] That is, the method of estimating the average classification cost (Section 2.2) is biased in favor of ICET (since ICET uses the method in its fitness function) and against the other algorithms. In this experiment, we labelled all tests as immediate. Otherwise, nothing changed from the baseline experiments. Table 7 summarizes the results of the experiment. ICET still performs well, although its advantage over the other algorithms has decreased slightly. Sensitivity to delayed tests is part of the explanation of ICET's performance, but it is not the whole story.

### 4.2.2 No Group Discounts

Another hypothesis is that ICET is superior simply because it can handle groups of tests that share a common cost. In this experiment, we eliminated group discounts for tests that share a common cost. That is, test costs were not conditional on prior tests. Otherwise, nothing changed from the baseline experiments. Table 8 summarizes the results of the experiment. ICET maintains its advantage over the other algorithms.

### 4.2.3 Complex Classification Cost Matrices

So far, we have only used simple classification cost matrices, where the penalty for a classification error is the same for all types of error. This assumption is not inherent in ICET. Each element in the classification cost matrix can have a different value. In this experiment, we explore ICET's behavior when the classification cost matrix is complex.

---

13. While the other algorithms cannot currently handle delayed tests, it should be possible to alter them in some way, so that they can handle delayed tests. This comment also extends to groups of tests that share a common cost. ICET might be viewed as an alteration of EG2 that enables EG2 to handle delayed tests and common costs.

Table 7: Average percentage of standard cost for the no-delay experiment.

| Algorithm | Average Classification Cost as Percentage of Standard ± 95% Confidence | |
|---|---|---|
| | Misclassification Error Costs from 10.00 to 10,000.00 | Misclassification Error Costs from 10.00 to 100.00 |
| ICET | 47 ± 6 | 28 ± 4 |
| EG2 | 54 ± 4 | 36 ± 2 |
| CS-ID3 | 54 ± 5 | 39 ± 3 |
| IDX | 54 ± 4 | 36 ± 2 |
| C4.5 | 64 ± 6 | 59 ± 4 |

Table 8: Average percentage of standard cost for the no-discount experiment.

| Algorithm | Average Classification Cost as Percentage of Standard ± 95% Confidence | |
|---|---|---|
| | Misclassification Error Costs from 10.00 to 10,000.00 | Misclassification Error Costs from 10.00 to 100.00 |
| ICET | 46 ± 6 | 25 ± 5 |
| EG2 | 56 ± 5 | 42 ± 3 |
| CS-ID3 | 59 ± 5 | 48 ± 4 |
| IDX | 56 ± 5 | 42 ± 3 |
| C4.5 | 75 ± 5 | 80 ± 4 |

We use the term "positive error" to refer to a false positive diagnosis, which occurs when a patient is diagnosed as being sick, but the patient is actually healthy. Conversely, the term "negative error" refers to a false negative diagnosis, which occurs when a patient is diagnosed as being healthy, but is actually sick. The term "positive error cost" is the cost that is assigned to positive errors, while "negative error cost" is the cost that is assigned to negative errors. See Appendix A for examples. We were interested in ICET's behavior as the ratio of negative to positive error cost was varied. Table 9 shows the ratios that we examined. Figure 5 shows the performance of the five algorithms at each ratio.

Our hypothesis was that the difference in performance between ICET and the other algorithms would increase as we move away from the middles of the plots, where the ratio is 1.0, since the other algorithms have no mechanism to deal with complex classification cost; they were designed under the implicit assumption of simple classification cost matrices. In fact, Figure 5 shows that the difference tends to *decrease* as we move away from the middles. This is most pronounced on the right-hand sides of the plots. When the ratio is 8.0 (the extreme right-hand sides of the plots), there is no advantage to using ICET. When the ratio is 0.125 (the extreme left-hand sides of the plots), there is still some advantage to using ICET.

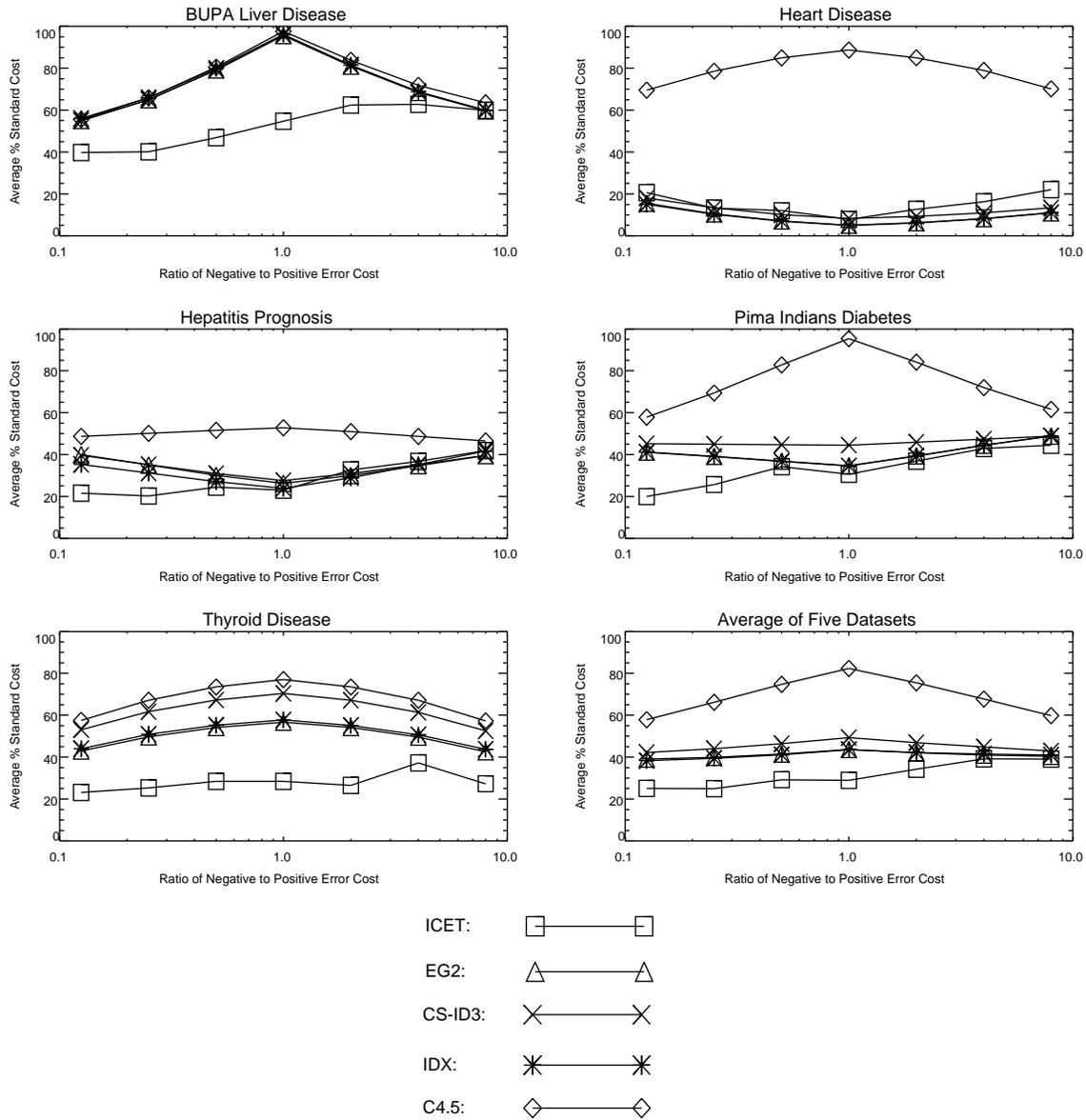

Figure 5: Average cost of classification as a percentage of the standard cost of classification, with complex classification cost matrices.

The interpretation of these plots is complicated by the fact that the gap between the algorithms tends to decrease as the penalty for classification errors increases (as we can see in Figure 3 — in retrospect, we should have held the sum of the negative error cost and the positive error cost at a constant value, as we varied their ratio). However, there is clearly an asymmetry in the plots, which we expected to be symmetrical about a vertical line centered on 1.0 on the x axis. The plots are close to symmetrical for the other algorithms, but they are asymmetrical for ICET. This is also apparent in Table 10, which focuses on a comparison of the performance of ICET and EG2, averaged across all five datasets (see the sixth plot in Figure 5). This suggests that it is more difficult to reduce negative errors (on the right-hand sides of the plots, negative errors have more weight) than it is to reduce positive errors (on the left-hand sides, positive errors have more weight). That is, it is easier to avoid false positive diagnoses (a patient is diagnosed as being sick, but the patient is actually healthy) than it is to avoid false negative diagnoses (a patient is diagnosed as being healthy, but is actually sick). This is unfortunate, since false negative diagnoses usually carry a heavier penalty, in real-life. Preliminary investigation suggests that false negative diagnoses are harder to avoid because the "sick" class is usually less frequent than the "healthy" class, which makes the "sick" class harder to learn.

Table 9: Actual error costs for each ratio of negative to positive error cost.

| Ratio of Negative to Positive Error Cost | Negative Error Cost | Positive Error Cost |
|---|---|---|
| 0.125 | 50 | 400 |
| 0.25 | 50 | 200 |
| 0.5 | 50 | 100 |
| 1.0 | 50 | 50 |
| 2.0 | 100 | 50 |
| 4.0 | 200 | 50 |
| 8.0 | 400 | 50 |

Table 10: Comparison of ICET and EG2
with various ratios of negative to positive error cost.

| Algorithm | Average Classification Cost as Percentage of Standard ± 95% Confidence, as the Ratio of Negative to Positive Error Cost is Varied | | | | | | |
|---|---|---|---|---|---|---|---|
| | 0.125 | 0.25 | 0.5 | 1.0 | 2.0 | 4.0 | 8.0 |
| ICET | 25 ± 10 | 25 ± 8 | 29 ± 6 | 29 ± 4 | 34 ± 6 | 39 ± 6 | 39 ± 6 |
| EG2 | 39 ± 5 | 40 ± 4 | 41 ± 4 | 44 ± 3 | 42 ± 3 | 41 ± 4 | 40 ± 5 |
| ICET/EG2 (as %) | 64 | 63 | 71 | 66 | 81 | 95 | 98 |

### 4.2.4 POORLY ESTIMATED CLASSIFICATION COST

We believe that it is an advantage of ICET that it is sensitive to both test costs and classification error costs. However, it might be argued that it is difficult to calculate the cost of classification errors in many real-world applications. Thus it is possible that an algorithm that ignores the cost of classification errors (e.g., EG2, CS-ID3, IDX) may be more robust and useful than an algorithm that is sensitive to classification errors (e.g., ICET). To address this possibility, we examine what happens when ICET is trained with a certain penalty for classification errors, then tested with a different penalty.

Our hypothesis was that ICET would be robust to reasonable differences between the penalty during training and the penalty during testing. Table 11 shows what happens when ICET is trained with a penalty of $100 for classification errors, then tested with penalties of

Table 11: Performance when training set classification error cost is $100.

| Algorithm | Average Classification Cost as Percentage of Standard ± 95% Confidence, for Testing Set Classification Error Cost of: | | |
| --- | --- | --- | --- |
| | $50 | $100 | $500 |
| ICET | 33 ± 10 | 41 ± 10 | 62 ± 9 |
| EG2 | 44 ± 3 | 49 ± 4 | 63 ± 6 |
| CS-ID3 | 49 ± 5 | 54 ± 6 | 65 ± 7 |
| IDX | 43 ± 3 | 49 ± 4 | 63 ± 6 |
| C4.5 | 82 ± 5 | 82 ± 5 | 78 ± 7 |

$50, $100, and $500. We see that ICET has the best performance of the five algorithms, although its edge is quite slight in the case where the penalty is $500 during testing.

We also examined what happens (1) when ICET is trained with a penalty of $500 and tested with penalties of $100, $500, and $1,000 and (2) when ICET is trained with a penalty of $1,000 and tested with penalties of $500, $1,000, and $5,000. The results show essentially the same pattern as in Table 11: ICET is relatively robust to differences between the training and testing penalties, at least when the penalties have the same order of magnitude. This suggests that ICET is applicable even in those situations where the reliability of the estimate of the cost of classification errors is dubious.

When the penalty for errors on the testing set is $100, ICET works best when the penalty for errors on the training set is also $100. When the penalty for errors on the testing set is $500, ICET works best when the penalty for errors on the training set is also $500. When the penalty for errors on the testing set is $1,000, ICET works best when the penalty for errors on the training set is $500. This suggests that there might be an advantage in some situations to underestimate the penalty for errors during training. In other, words ICET may have a tendency to overestimate the benefits of tests (this is likely due to overfitting the training data).

## 4.3 Searching Bias Space

The final group of experiments analyzes ICET's method for searching in bias space. Section 4.3.1 studies the roles of the mutation and crossover operators. It appears that crossover is mildly beneficial, compared to pure mutation. Section 4.3.2 considers what happens when ICET is constrained to search in a binary bias space, instead of a real bias space. This constraint actually improves the performance of ICET. We hypothesized that the improvement was due to a hidden advantage of searching in binary bias space: When searching in binary bias space, ICET has direct access to the true costs of the tests. However, this advantage can be available when searching in real bias space, if the initial population of biases is seeded with the true costs of the tests. Section 4.3.3 shows that this seeding improves the performance of ICET.

### 4.3.1 Crossover Versus Mutation

Past work has shown that a genetic algorithm with crossover performs better than a genetic algorithm with mutation alone (Grefenstette *et al.*, 1990; Wilson, 1987). This section attempts to test the hypothesis that crossover improves the performance of ICET. To test this hypothesis, it is not sufficient to merely set the crossover rate to zero. Since crossover has a randomizing effect, similar to mutation, we must also increase the mutation rate, to compensate for the loss of crossover (Wilson, 1987; Spears, 1992).

It is very difficult to analytically calculate the increase in mutation rate that is required to compensate for the loss of crossover (Spears, 1992). Therefore we experimentally tested three different mutation settings.[14] The results are summarized in Table 12. When the crossover rate was set to zero, the best mutation rate was 0.10. For misclassification error costs from $10 to $10,000, the performance of ICET without crossover was not as good as the performance of ICET with crossover, but the difference is not statistically significant. However, this comparison is not entirely fair to crossover, since we made no attempt to optimize the crossover rate (we simply used the default value). The results suggest that crossover is mildly beneficial, but do not prove that pure mutation is inferior.

Table 12: Average percentage of standard cost for mutation experiment.

| ICET | | Average Classification Cost as Percentage of Standard ± 95% Confidence | |
|---|---|---|---|
| Crossover Rate | Mutation Rate | Misclassification Error Costs from 10.00 to 10,000.00 | Misclassification Error Costs from 10.00 to 100.00 |
| 0.6 | 0.001 | 49 ± 7 | 29 ± 7 |
| 0.0 | 0.05 | 51 ± 8 | 32 ± 9 |
| 0.0 | 0.10 | 50 ± 8 | 29 ± 8 |
| 0.0 | 0.15 | 51 ± 8 | 30 ± 9 |

### 4.3.2 SEARCH IN BINARY SPACE

ICET searches for biases in a space of $n + 2$ real numbers. Inspired by Aha and Bankert (1994), we decided to see what would happen when ICET was restricted to a space of $n$ binary numbers and 2 real numbers. We modified ICET so that EG2 was given the true cost of each test, instead of a "pseudo-cost" or bias. For conditional test costs, we used the no-discount cost (see Section 4.2.2). The $n$ binary digits were used to exclude or include a test. EG2 was not allowed to use excluded tests in the decision trees that it generated.

To be more precise, let $B_1, ..., B_n$ be $n$ binary numbers and let $C_1, ..., C_n$ be $n$ real numbers. For this experiment, we set $C_i$ to the true cost of the $i$-th test. In this experiment, GENESIS does not change $C_i$. That is, $C_i$ is constant for a given test in a given dataset. Instead, GENESIS manipulates the value of $B_i$ for each $i$. The binary number $B_i$ is used to determine whether EG2 is allowed to use a test in its decision tree. If $B_i = 0$, then EG2 is not allowed to use the $i$-th test (the $i$-th attribute). Otherwise, if $B_i = 1$, EG2 is allowed to use the $i$-th test. EG2 uses the ICF equation as usual, with the true costs $C_i$. Thus this modified version of ICET is searching through a binary bias space instead of a real bias space.

Our hypothesis was that ICET would perform better when searching in real bias space than when searching in binary bias space. Table 13 shows that this hypothesis was not confirmed. It appears to be better to search in binary bias space, rather than real bias space. However, the differences are not statistically significant.

---

14. Each of these three experiments took one week on a Sparc 10, which is why we only tried three settings for the mutation rate.

Table 13: Average percentage of standard cost for the binary search experiment.

| Algorithm | Average Classification Cost as Percentage of Standard ± 95% Confidence | |
| | Misclassification Error Costs from 10.00 to 10,000.00 | Misclassification Error Costs from 10.00 to 100.00 |
| --- | --- | --- |
| ICET — Binary Space | 48 ± 6 | 26 ± 5 |
| ICET — Real Space | 49 ± 7 | 29 ± 7 |
| EG2 | 58 ± 5 | 43 ± 3 |
| CS-ID3 | 61 ± 6 | 49 ± 4 |
| IDX | 58 ± 5 | 43 ± 3 |
| C4.5 | 77 ± 5 | 82 ± 4 |

When we searched in binary space, we set $C_i$ to the true cost of the $i$-th test. GENESIS manipulated $B_i$ instead of $C_i$. When we searched in real space, GENESIS set $C_i$ to whatever value it found useful in its attempt to optimize fitness. We hypothesized that this gives an advantage to binary space search over real space search. Binary space search has direct access to the true costs of the tests, but real space search only learns about the true costs of the tests indirectly, by the feedback it gets from the fitness function.

When we examined the experiment in detail, we found that ICET did well on the Heart Disease dataset when it was searching in binary bias space, although it did poorly when it was searching in real bias space (see Section 4.1.1). We hypothesized that ICET, when searching in real space, suffered most from the lack of direct access to the true costs when it was applied to the Heart Disease dataset. These hypotheses were tested by the next experiment.

### 4.3.3 Seeded Population

In this experiment, we returned to searching in real bias space, but we seeded the initial population of biases with the true test costs. This gave ICET direct access to the true costs. For conditional test costs, we used the no-discount cost (see Section 4.2.2). In the baseline experiment (Section 4.1), the initial population consists of 50 randomly generated strings, representing $n + 2$ real numbers. In this experiment, the initial population consists of 49 randomly generated strings and one manually generated string. In the manually generated string, the first $n$ numbers are the true test costs. The last two numbers were set to 1.0 (for ω) and 25 (for CF). This string is exactly the bias of EG2, as implemented here (Section 3.2).

Our hypotheses were (1) that ICET would perform better (on average) when the initial population is seeded than when it is purely random, (2) that ICET would perform better (on average) searching in real space with a seeded population than when searching in binary space,[15] and (3) that ICET would perform better on the Heart Disease dataset when the ini-tial population is seeded than when it is purely random. Table 14 appears to support the first two hypotheses. Figure 6 appears to support the third hypothesis. However, the results are not statistically significant.[16]

Table 14: Average percentage of standard cost for the seeded population experiment.

| Algorithm | Average Classification Cost as Percentage of Standard ± 95% Confidence | |
| | Misclassification Error Costs from 10.00 to 10,000.00 | Misclassification Error Costs from 10.00 to 100.00 |
| --- | --- | --- |
| ICET — Seeded Search in Real Space | 46 ± 6 | 25 ± 5 |
| ICET — Unseeded Search in Real Space | 49 ± 7 | 29 ± 7 |
| ICET — Unseeded Search in Binary Space | 48 ± 6 | 26 ± 5 |
| EG2 | 58 ± 5 | 43 ± 3 |
| CS-ID3 | 61 ± 6 | 49 ± 4 |
| IDX | 58 ± 5 | 43 ± 3 |
| C4.5 | 77 ± 5 | 82 ± 4 |

This experiment raises some interesting questions: Should seeding the population be built into the ICET algorithm? Should we seed the whole population with the true costs, perturbed by some random noise? Perhaps this is the right approach, but we prefer to modify $ICF_i$ (equation (2)), the device by which GENESIS controls the decision tree induction. We could alter this equation so that it contains both the true costs and some bias parameters.[17] This seems to make more sense than our current approach, which deprives EG2 of direct access to the true costs. We discuss some other ideas for modifying the equation in Section 5.2.

Incidentally, this experiment lets us answer the following question: Does the genetic search in bias space do anything useful? If we start with the true costs of the tests and reasonable values for the parameters ω and CF, how much improvement do we get from the genetic search? In this experiment, we seeded the population with an individual that represents exactly the bias of EG2 (the first $n$ numbers are the true test costs and the last two numbers are 1.0 for ω and 25 for CF). Therefore we can determine the value of genetic search by comparing EG2 with ICET. ICET starts with the bias of EG2 (as a seed in the first genera-tion) and attempts to improve the bias. The score of EG2 in Table 14 shows the value of the bias built into EG2. The score of ICET in Table 14 shows how genetic search in bias space can improve the built-in bias of EG2. When the cost of misclassification errors has the same order of magnitude as the test costs ($10 to $100), EG2 averages 43% of the standard cost, while ICET averages 25% of the standard cost. When the cost of misclassification errors ranges from $10 to $10,000, EG2 averages 58% of the standard cost, while ICET averages 46% of the standard cost. Both of these differences are significant with more than 95% confidence. This makes it clear that genetic search is adding value.

---

15. Note that it does not make sense to seed the binary space search, since it already has direct access to the true costs.

16. We would need to go from the current 10 trials (10 random splits of the data) to about 40 trials to make the results significant. The experiments reported here took a total of 63 days of continuous computation on a Sun Sparc 10, so 40 trials would require about six more months.

17. This idea was suggested in conversation by K. De Jong.

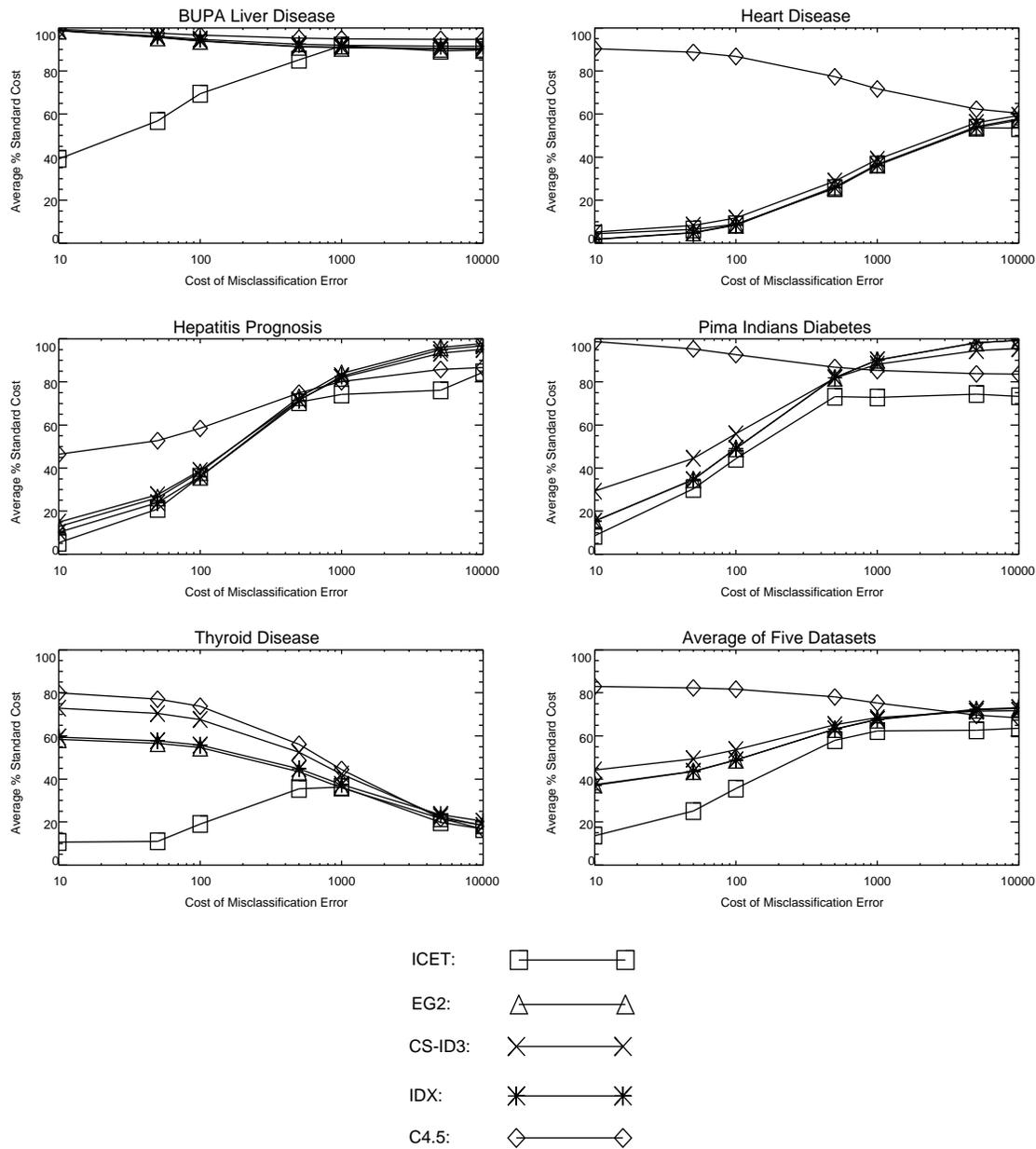

Figure 6: Average cost of classification as a percentage of the standard cost of classification for the seeded population experiment.

## 5. Discussion

This section compares ICET to related work and outlines some possibilities for future work.

### 5.1 Related Work

There are several other algorithms that are sensitive to test costs (Núñez, 1988, 1991; Tan & Schlimmer, 1989, 1990; Tan, 1993; Norton, 1989). As we have discussed, the main limitation of these algorithms is that they do not consider the cost of classification errors. We cannot rationally determine whether a test should be performed until we know both the cost of the test and the cost of classification errors.

There are also several algorithms that are sensitive to classification error costs (Breiman *et al.*, 1984; Friedman & Stuetzle, 1981; Hermans *et al.*, 1974; Gordon & Perlis, 1989; Pazzani *et al.*, 1994; Provost, 1994; Provost & Buchanan, in press; Knoll *et al.*, 1994). None of these algorithms consider the cost of tests. Therefore they all focus on complex classification cost matrices, since, when tests have no cost and the classification error matrix is simple, the problem reduces to maximizing accuracy.

The FIS system (Pipitone *et al.*, 1991) attempts to find a decision tree that minimizes the average total cost of the tests required to achieve a certain level of accuracy. This approach could be implemented in ICET by altering the fitness function. The main distinction between FIS (Pipitone *et al.*, 1991) and ICET is that FIS does not learn from data. The information gain of a test is estimated using a *qualitative causal model*, instead of training cases. Qualitative causal models are elicited from domain experts, using a special knowledge acquisition tool. When training data are available, ICET can be used to avoid the need for knowledge acquisition. Otherwise, ICET is not applicable and the FIS approach is suitable.

Another feature of ICET is that it does not perform purely greedy search. Several other authors have proposed non-greedy classification algorithms (Tcheng *et al.*, 1989; Ragavan & Rendell, 1993; Norton, 1989; Schaffer, 1993; Rymon, 1993; Seshu, 1989). In general, these results show that there can be an advantage to more sophisticated search procedures. ICET is different from these algorithms in that it uses a genetic algorithm and it is applied to minimizing both test costs and classification error costs.

ICET uses a two-tiered search strategy. At the bottom tier, EG2 performs a greedy search through the space of classifiers. On the second tier, GENESIS performs a non-greedy search through a space of biases. The idea of a two-tiered search strategy (where the first tier is search in classifier space and the second tier is search in bias space) also appears in (Provost, 1994; Provost & Buchanan, in press; Aha & Bankert, 1994; Schaffer, 1993). Our work goes beyond Aha and Bankert (1994) by considering search in a real bias space, rather than search in a binary space. Our work fits in the general framework of Provost and Buchanan (in press), but differs in many details. For example, their method of calculating cost is a special case of ours (Section 2.3).

Other researchers have applied genetic algorithms to classification problems. For example, Frey and Slate (1991) applied a genetic algorithm (in particular, a learning classifier system (LCS)) to letter recognition. However, Fogarty (1992) obtained higher accuracy using a simple nearest neighbor algorithm. More recent applications of genetic algorithms to classification have been more successful (De Jong *et al.*, 1993). However, the work described here is the first application of genetic algorithms to the problem of cost-sensitive classification.

We mentioned in Section 2.1 that decision theory may be used to define the optimal solution to the problem of cost-sensitive classification. However, searching for the optimal solution is computationally infeasible (Pearl, 1988). We attempted to take a decision theoretic approach to this problem by implementing the AO* algorithm (Pearl, 1984) and designing a heuristic evaluation function to speed up the AO* search (Lirov & Yue, 1991). We were unable to make this approach execute fast enough to be practical.

We also attempted to apply genetic programming (Koza, 1993) to the problem of cost-sensitive classification. Again, we were unable to make this approach execute fast enough to be practical, although it was faster than the AO* approach.

The cost-sensitive classification problem, as we have treated it here, is essentially a problem in reinforcement learning (Sutton, 1992; Karakoulas, in preparation). The average cost of classification, measured as described in Section 2.2, is a reward/punishment signal that could be optimized using reinforcement learning techniques. This is something that might be explored as an alternative approach.

## 5.2  Future Work

This paper discusses two types of costs, the cost of tests and the cost of misclassification errors. These two costs have been treated together in decision theory, but ICET is the first machine learning system that handles both costs together. The experiments in this paper have compared ICET to other machine learning systems that can handle test costs (Núñez, 1988, 1991; Tan & Schlimmer, 1989, 1990; Tan, 1993; Norton, 1989), but we have not compared ICET to other machine learning systems that can handle classification error costs (Breiman *et al.*, 1984; Friedman & Stuetzle, 1981; Hermans *et al.*, 1974; Gordon & Perlis, 1989; Pazzani *et al.*, 1994; Provost, 1994; Provost & Buchanan, in press; Knoll *et al.*, 1994). In future work, we plan to address this omission. A proper treatment of this issue would make this paper too long.

The absence of comparison with machine learning systems that can handle classification error costs has no impact on most of the experiments reported here. The experiments in this paper focussed on simple classification cost matrices (except for Section 4.2.3). When the classification cost matrix is simple and the cost of tests is ignored, minimizing the cost is exactly equivalent to maximizing accuracy (see Section 2.3). Therefore, C4.5 (which is designed to maximize accuracy) is a suitable surrogate for any of the systems that can handle classification error costs.

We also did not experiment with setting the test costs to zero. However, the behavior of ICET when the penalty for misclassification errors is very high (the extreme right-hand sides of the plots in Figure 3) is necessarily the same as its behavior when the cost of tests is very low, since ICET is sensitive to the relative differences between test costs and error costs, not the absolute costs. Therefore (given the behavior we can observe in the extreme right-hand sides of the plots in Figure 3) we can expect that the performance of ICET will tend to converge with the performance of the other algorithms as the cost of tests approaches zero.

One natural addition to ICET would be the ability to output an "I don't know" class. This is easily handled by the GENESIS component, by extending the classification cost matrix so that a cost is assigned to classifying a case as "unknown". We need to also make a small modification to the EG2 component, so that it can generate decision trees with leaves labelled "unknown". One way to do this would be to introduce a parameter that defines a confidence threshold. Whenever the confidence in a certain leaf drops below the confidence threshold, that leaf would be labelled "unknown". This confidence parameter would be made accessible to the GENESIS component, so that it could be tuned to minimize average classification cost.

The mechanism in ICET for handling conditional test costs has some limitations. As it is currently implemented, it does not handle the cost of attributes that are calculated from other attributes. For example, in the Thyroid dataset (Appendix A.5), the FTI test is calculated based on the results of the TT4 and T4U tests. If the FTI test is selected, we must pay for the TT4 and T4U tests. If the TT4 and T4U tests have already been selected, the FTI test is free (since the calculation is trivial). The ability to deal with calculated test results could be added to ICET with relatively little effort.

ICET, as currently implemented, only handles two classes of test results: tests with "immediate" results and tests with "delayed" results. Clearly there can be a continuous range of delays, from seconds to years. We have chosen to treat delays as distinct from test costs, but it could be argued that a delay is simply another type of test cost. For example, we could say that a group of blood tests shares the common cost of a one-day wait for results. The cost of one of the blood tests is conditional on whether we are prepared to commit ourselves to doing one or more of the other tests in the group, before we see the results of the first test. One difficulty with this approach to handling delays is the problem of assigning a cost to the delay. How much does it cost to bring a patient in for two blood samples, instead of one? Do we include the disruption to the patient's life in our estimate of the cost? To avoid these questions, we have not treated delays as another type of test cost, but our approach does not readily handle a continuous range of delays.

The cost of a test can be a function of several things: (1) It can be a function of the prior tests that have been selected. (2) It can be a function of the actual class of the case. (3) It can be a function of other aspects of the case, where information about these other aspects may be available through other tests. (4) It can be a function of the test result. This list seems comprehensive, but there may be some possibilities we have overlooked. Let us consider each of these four possibilities.

First, the cost of a test can be a function of the prior tests that have been selected. ICET handles a special case of this, where a group of tests shares a common cost. As it is currently implemented, ICET does not handle the general case. However, we could easily add this capability to ICET by modifying the fitness function.

Second, the cost of a test can be a function of the actual class of the case. For example, a test for heart disease might involve heavy exercise (Appendix A.2). If the patient actually has heart disease, the exercise might trigger a heart attack. This risk should be included in the cost of this particular test. Thus the cost of this test should vary, depending on whether the patient actually has heart disease. We have not implemented this, although it could easily be added to ICET by modifying the fitness function.

Third, the cost of a test can be a function of the results of other tests. For example, drawing blood from a newborn is more costly than drawing blood from an adult. To assign a cost to a blood test, we need to know the age of the patient. The age of the patient can be represented as the result of another test — the "patient-age" test. This is slightly more complex than the preceding cases, because we must now insure that the blood test is always accompanied with the patient-age test. We have not implemented this, although it could be added to ICET.

Fourth, the cost of a test can be a function of the test result. For example, injecting a radio-opaque die for an X-ray might cause an allergic reaction in the patient. This risk should be added to the cost of the test. This makes the cost of the test a function of one of the possible outcomes of the test. In a situation like this, it may be wise to precede the injection of the die with a screening test for allergies. This could be as simple as asking a question to the patient. This question may have no relevance at all for determining the correct diagnosis of the patient, but it may serve to reduce the average cost of classification. This case is similar to the third case, above. Again, we have not implemented this, although it could be added to ICET.

Attribute selection in EG2, CS-ID3, and IDX shares a common form. We may view attribute selection as a function from $\Re^n$ to $\{1, ..., n\}$, which takes as input $n$ information gain values $\Delta I_1, ..., \Delta I_n$ (one for each attribute) and generates as output the index of one of the attributes. We may view $C_1, ..., C_n$ and $\omega$ as parameters in the attribute selection function. These parameters may be used to control the bias of the attribute selection procedure. In this view, ICET uses GENESIS to tune the parameters of EG2's attribute selection function. In the future, we would like to investigate more general attribute selection functions. For example, we might use a neural network to implement a function from $\Re^n$ to $\{1, ..., n\}$. GENESIS would then be used to tune the weights in the neural network.[18] The attribute selection function might also benefit from the addition of an input that specifies the depth of the decision tree at the current node, where the information gain values are measured. This would enable the bias for a test to vary, depending on how many tests have already been selected.

Another area for future work is to explore the parameter settings that control GENESIS (Table 4). There are many parameters that could be adjusted in GENESIS. We think it is significant that ICET works well with the default parameter settings in GENESIS, since it shows that ICET is robust with respect to the parameters. However, it might be possible to substantially improve the performance of ICET by tuning some of these parameters. A recent trend in genetic algorithm research is to let the genetic algorithm adjust some of its own parameters, such as mutation rate and crossover rate (Whitley *et al.*, 1993). Another possibility is to stop breeding when the fitness levels stop improving, instead of stopping after a fixed number of generations. Provost and Buchanan (in press) use a goodness measure as a stopping condition for the bias space search.

## 6. Conclusions

The central problem investigated here is the problem of minimizing the cost of classification when the tests are expensive. We argued that this requires assigning a cost to classification errors. We also argued that a decision tree is the natural form of knowledge representation for this type of problem. We then presented a general method for calculating the average cost of classification for a decision tree, given a decision tree, information on the calculation of test costs, a classification cost matrix, and a set of testing data. This method is applicable to standard classification decision trees, without regard to how the decision tree is generated. The method is sensitive to test costs, sensitive to classification error costs, capable of handling conditional test costs, and capable of handling delayed tests.

We introduced ICET, a hybrid genetic decision tree induction algorithm. ICET uses a genetic algorithm to evolve a population of biases for a decision tree induction algorithm. Each individual in the population represents one set of biases. The fitness of an individual is determined by using it to generate a decision tree with a training dataset, then calculating the average cost of classification for the decision tree with a testing dataset.

We analyzed the behavior of ICET in a series of experiments, using five real-world medical datasets. Three groups of experiments were performed. The first group looked at the baseline performance of the five algorithms on the five datasets. ICET was found to have sig-nificantly lower costs than the other algorithms. Although it executes more slowly, an average time of 23 minutes (for a typical dataset) is acceptable for many applications, and there is the possibility of much greater speed on a parallel machine. The second group of experiments studied the robustness of ICET under a variety of modifications to its input. The results show that ICET is robust. The third group of experiments examined ICET's search in bias space. We discovered that the search could be improved by seeding the initial population of biases.

---

18. This idea was suggested in conversation by M. Brooks.

In general, our research is concerned with pragmatic constraints on classification problems (Provost & Buchanan, in press). We believe that many real-world classification problems involve more than merely maximizing accuracy (Turney, in press). The results presented here indicate that, in certain applications, a decision tree that merely maximizes accuracy (e.g., trees generated by C4.5) may be far from the performance that is possible with an algorithm that considers such realistic constraints as test costs, classification error costs, conditional test costs, and delayed test results. These are just a few of the pragmatic constraints that are faced in real-world classification problems.

## Appendix A. Five Medical Datasets

This appendix presents the test costs for five medical datasets, taken from the Irvine collection (Murphy & Aha, 1994). The costs are based on information from the Ontario Ministry of Health (1992). Although none of the medical data were gathered in Ontario, it is reasonable to assume that other areas have similar relative test costs. For our purposes, the relative costs are important, not the absolute costs.

### A.1 BUPA Liver Disorders

The BUPA Liver Disorders dataset was created by BUPA Medical Research Ltd. and it was donated to the Irvine collection by Richard Forsyth.[19] Table 15 shows the test costs for the BUPA Liver Disorders dataset. The tests in group A are blood tests that are thought to be sensitive to liver disorders that might arise from excessive alcohol consumption. These tests share the common cost of $2.10 for collecting blood. The target concept was defined using the sixth column: Class 0 was defined as "drinks < 3" and class 1 was defined as "drinks ≥ 3". Table 16 shows the general form of the classification cost matrix that was used in the experiments in Section 4. For most of the experiments, the classification error cost equals the positive error cost equals the negative error cost. The exception is in Section 4.2.3, for the experiments with complex classification cost matrices. The terms "positive error cost" and "negative error cost" are explained in Section 4.2.3. There are 345 cases in this dataset, with no missing values. Column seven was originally used to split the data into training and testing sets. We did not use this column, since we required ten different random splits of the data. In our ten random splits, the ten training sets all had 230 cases and the ten testing sets all had 115 cases.

### A.2 Heart Disease

The Heart Disease dataset was donated to the Irvine collection by David Aha.[20] The princi-

Table 15: Test costs for the BUPA Liver Disorders dataset.

|   | Test | Description | Group | Cost | Delayed |
|---|------|-------------|-------|------|---------|
| 1 | mcv | mean corpuscular volume | A | $7.27 if first test in group A, $5.17 otherwise | yes |
| 2 | alkphos | alkaline phosphotase | A | $7.27 if first test in group A, $5.17 otherwise | yes |
| 3 | sgpt | alamine aminotransferase | A | $7.27 if first test in group A, $5.17 otherwise | yes |
| 4 | sgot | aspartate aminotransferase | A | $7.27 if first test in group A, $5.17 otherwise | yes |
| 5 | gammagt | gamma-glutamyl transpeptidase | A | $9.86 if first test in group A, $7.76 otherwise | yes |
| 6 | drinks | number of half-pint equivalents of alcoholic beverages drunk per day | | diagnostic class: "drinks < 3" or "drinks ≥ 3" | - |
| 7 | selector | field used to split data into two sets | | not used | - |

Table 16: Classification costs for the BUPA Liver Disorders dataset.

| Actual Class | Guess Class | Cost |
|--------------|-------------|------|
| 0 (drinks < 3) | 0 (drinks < 3) | $0.00 |
| 0 (drinks < 3) | 1 (drinks ≥ 3) | Positive Error Cost |
| 1 (drinks ≥ 3) | 0 (drinks < 3) | Negative Error Cost |
| 1 (drinks ≥ 3) | 1 (drinks ≥ 3) | $0.00 |

pal medical investigator was Robert Detrano, of the Cleveland Clinic Foundation. Table 17 shows the test costs for the Heart Disease dataset. A nominal cost of $1.00 was assigned to the first four tests. The tests in group A are blood tests that are thought to be relevant for heart disease. These tests share the common cost of $2.10 for collecting blood. The tests in groups B and C involve measurements of the heart during exercise. A nominal cost of $1.00 was assigned for tests after the first test in each of these groups. The class variable has the values "buff" (healthy) and "sick". There was a fifteenth column, which specified the class variable as "H" (healthy), "S1", "S2", "S3", or "S4" (four different types of "sick"), but we deleted this column. Table 18 shows the classification cost matrix. There are 303 cases in this dataset. We deleted all cases for which there were missing values. This reduced the dataset to 296 cases. In our ten random splits, the training sets had 197 cases and the testing sets had 99 cases.

### A.3    Hepatitis Prognosis

The Hepatitis Prognosis dataset was donated by Gail Gong.[21] Table 19 shows the test costs for the Hepatitis dataset. Unlike the other four datasets, this dataset deals with prognosis, not diagnosis. With prognosis, the diagnosis is known, and the problem is to determine the likely outcome of the disease. The tests that were assigned a nominal cost of $1.00 either involve asking a question to the patient or performing a basic physical examination on the patient. The tests in group A share the cost of $2.10 for collecting blood. Note that, although performing a histological examination of the liver costs $81.64, asking the patient whether a histology was performed only costs $1.00. Thus the prognosis can exploit the information conveyed by a decision (to perform a histological examination) that was made during the diagnosis. The class variable has the values 1 (die) and 2 (live). Table 20 shows the classification costs. The dataset contains 155 cases, with many missing values. In our ten random splits, the training sets had 103 cases and the testing sets had 52 cases. We filled in the missing values, using a simple single nearest neighbor algorithm (Aha *et al.*, 1991). The missing values were filled in using the whole dataset, before the dataset was split into training and testing sets. For the nearest neighbor algorithm, the data were normalized so that the mini-mum value of a feature was 0 and the maximum value was 1. The distance measure used was the sum of the absolute values of the differences. The difference between two values was defined to be 1 if one or both of the two values was missing.

---

21. The Hepatitis Prognosis dataset has the URL ftp://ftp.ics.uci.edu/pub/machine-learning-databases/hepatitis/hepatitis.data.

Table 17: Test costs for the Heart Disease dataset.

| | Test | Description | Group | Cost | Delayed |
|---|---|---|---|---|---|
| 1 | age | age in years | | $1.00 | no |
| 2 | sex | patient's gender | | $1.00 | no |
| 3 | cp | chest pain type | | $1.00 | no |
| 4 | trestbps | resting blood pressure | | $1.00 | no |
| 5 | chol | serum cholesterol | A | $7.27 if first test in group A, $5.17 otherwise | yes |
| 6 | fbs | fasting blood sugar | A | $5.20 if first test in group A, $3.10 otherwise | yes |
| 7 | restecg | resting electrocardiograph | | $15.50 | yes |
| 8 | thalach | maximum heart rate achieved | B | $102.90 if first test in group B, $1.00 otherwise | yes |
| 9 | exang | exercise induced angina | C | $87.30 if first test in group C, $1.00 otherwise | yes |
| 10 | oldpeak | ST depression induced by exercise relative to rest | C | $87.30 if first test in group C, $1.00 otherwise | yes |
| 11 | slope | slope of peak exercise ST segment | C | $87.30 if first test in group C, $1.00 otherwise | yes |
| 12 | ca | number of major vessels coloured by fluoroscopy | | $100.90 | yes |
| 13 | thal | 3 = normal; 6 = fixed defect; 7 = reversible defect | B | $102.90 if first test in group B, $1.00 otherwise | yes |
| 14 | num | diagnosis of heart disease | | diagnostic class | - |

Table 18: Classification costs for the Heart Disease dataset.

| Actual Class | Guess Class | Cost |
|---|---|---|
| buff | buff | $0.00 |
| buff | sick | Positive Error Cost |
| sick | buff | Negative Error Cost |
| sick | sick | $0.00 |

Table 19: Test costs for the Hepatitis Prognosis dataset.

|    | Test           | Description                   | Group | Cost                                              | Delayed |
|----|----------------|------------------------------|-------|---------------------------------------------------|---------|
| 1  | class          | prognosis of hepatitis       |       | prognostic class: live or die                     | -       |
| 2  | age            | age in years                 |       | $1.00                                             | no      |
| 3  | sex            | gender                       |       | $1.00                                             | no      |
| 4  | steroid        | patient on steroids          |       | $1.00                                             | no      |
| 5  | antiviral      | patient on antiviral         |       | $1.00                                             | no      |
| 6  | fatigue        | patient reports fatigue      |       | $1.00                                             | no      |
| 7  | malaise        | patient reports malaise      |       | $1.00                                             | no      |
| 8  | anorexia       | patient anorexic             |       | $1.00                                             | no      |
| 9  | liver big      | liver big on physical exam   |       | $1.00                                             | no      |
| 10 | liver firm     | liver firm on physical exam  |       | $1.00                                             | no      |
| 11 | spleen palpable| spleen palpable on physical  |       | $1.00                                             | no      |
| 12 | spiders        | spider veins visible         |       | $1.00                                             | no      |
| 13 | ascites        | ascites visible              |       | $1.00                                             | no      |
| 14 | varices        | varices visible              |       | $1.00                                             | no      |
| 15 | bilirubin      | bilirubin — blood test       | A     | $7.27 if first test in group A, $5.17 otherwise   | yes     |
| 16 | alk phosphate  | alkaline phosphotase         | A     | $7.27 if first test in group A, $5.17 otherwise   | yes     |
| 17 | sgot           | aspartate aminotransferase   | A     | $7.27 if first test in group A, $5.17 otherwise   | yes     |
| 18 | albumin        | albumin — blood test         | A     | $7.27 if first test in group A, $5.17 otherwise   | yes     |
| 19 | protime        | protime — blood test         | A     | $8.30 if first test in group A, $6.20 otherwise   | yes     |
| 20 | histology      | was histology performed?     |       | $1.00                                             | no      |

Table 20: Classification costs for the Hepatitis Prognosis dataset.

| Actual Class | Guess Class | Cost                |
|--------------|-------------|---------------------|
| 1 (die)      | 1 (die)     | $0.00               |
| 1 (die)      | 2 (live)    | Negative Error Cost |
| 2 (live)     | 1 (die)     | Positive Error Cost |
| 2 (live)     | 2 (live)    | $0.00               |

### A.4 Pima Indians Diabetes

The Pima Indians Diabetes dataset was donated by Vincent Sigillito.[22] The data were collected by the National Institute of Diabetes and Digestive and Kidney Diseases. Table 21 shows the test costs for the Pima Indians Diabetes dataset. The tests in group A share the cost of $2.10 for collecting blood. The remaining tests were assigned a nominal cost of $1.00. All of the patients were females at least 21 years old of Pima Indian heritage. The class variable has the values 0 (healthy) and 1 (diabetes). Table 22 shows classification costs. The dataset includes 768 cases, with no missing values. In our ten random splits, the training sets had 512 cases and the testing sets had 256 cases.

Table 21: Test costs for the Pima Indians Diabetes dataset.

|   | Test | Description | Group | Cost | Delayed |
|---|------|-------------|-------|------|---------|
| 1 | times pregnant | number of times pregnant | | $1.00 | no |
| 2 | glucose tol | glucose tolerance test | A | $17.61 if first test in group A, $15.51 otherwise | yes |
| 3 | diastolic bp | diastolic blood pressure | | $1.00 | no |
| 4 | triceps | triceps skin fold thickness | | $1.00 | no |
| 5 | insulin | serum insulin test | A | $22.78 if first test in group A, $20.68 otherwise | yes |
| 6 | mass index | body mass index | | $1.00 | no |
| 7 | pedigree | diabetes pedigree function | | $1.00 | no |
| 8 | age | age in years | | $1.00 | no |
| 9 | class | diagnostic class | | diagnostic class | - |

Table 22: Classification costs for the Pima Indians Diabetes dataset.

| Actual Class | Guess Class | Cost |
|--------------|-------------|------|
| 0 (healthy) | 0 (healthy) | $0.00 |
| 0 (healthy) | 1 (diabetes) | Positive Error Cost |
| 1 (diabetes) | 0 (healthy) | Negative Error Cost |
| 1 (diabetes) | 1 (diabetes) | $0.00 |

### A.5 Thyroid Disease

The Thyroid Disease dataset was created by the Garavan Institute, Sydney, Australia. The file was donated by Randolf Werner, obtained from Daimler-Benz. Daimler-Benz obtained the data from J.R. Quinlan.[23] Table 23 shows the test costs for the Thyroid Disease dataset. A nominal cost of $1.00 was assigned to the first 16 tests. The tests in group A share the cost of $2.10 for collecting blood. The FTI test involves a calculation based on the results of the TT4 and T4U tests. This complicates the calculation of the costs of these three tests, so we chose not to use the FTI test in our experiments. The class variable has the values 1 (hypothyroid), 2 (hyperthyroid), and 3 (normal). Table 24 shows the classification costs. There are 3772 cases in this dataset, with no missing values. In our ten random splits, the training sets had 2515 cases and the testing sets had 1257 cases.

---

22. The Pima Indians Diabetes dataset has the URL ftp://ftp.ics.uci.edu/pub/machine-learning-databases/pima-indians-diabetes/pima-indians-diabetes.data.

Table 23: Test costs for the Thyroid Disease dataset.

| | Test | Description | Group | Cost | Delayed |
|---|---|---|---|---|---|
| 1 | age | age in years | | $1.00 | no |
| 2 | sex | gender | | $1.00 | no |
| 3 | on thyroxine | patient on thyroxine | | $1.00 | no |
| 4 | query thyroxine | maybe on thyroxine | | $1.00 | no |
| 5 | on antithyroid | on antithyroid medication | | $1.00 | no |
| 6 | sick | patient reports malaise | | $1.00 | no |
| 7 | pregnant | patient pregnant | | $1.00 | no |
| 8 | thyroid surgery | history of thyroid surgery | | $1.00 | no |
| 9 | I131 treatment | patient on I131 treatment | | $1.00 | no |
| 10 | query hypothyroid | maybe hypothyroid | | $1.00 | no |
| 11 | query hyperthyroid | maybe hyperthyroid | | $1.00 | no |
| 12 | lithium | patient on lithium | | $1.00 | no |
| 13 | goitre | patient has goitre | | $1.00 | no |
| 14 | tumour | patient has tumour | | $1.00 | no |
| 15 | hypopituitary | patient hypopituitary | | $1.00 | no |
| 16 | psych | psychological symptoms | | $1.00 | no |
| 17 | TSH | TSH value, if measured | A | $22.78 if first test in group A, $20.68 otherwise | yes |
| 18 | T3 | T3 value, if measured | A | $11.41 if first test in group A, $9.31 otherwise | yes |
| 19 | TT4 | TT4 value, if measured | A | $14.51 if first test in group A, $12.41 otherwise | yes |
| 20 | T4U | T4U value, if measured | A | $11.41 if first test in group A, $9.31 otherwise | yes |
| 21 | FTI | FTI — calculated from TT4 and T4U | | not used | - |
| 22 | class | diagnostic class | | diagnostic class | - |

Table 24: Classification costs for the Thyroid Disease dataset.

| Actual Class | Guess Class | Cost |
|---|---|---|
| 1 (hypothyroid) | 1 (hypothyroid) | $0.00 |
| 1 (hypothyroid) | 2 (hyperthyroid) | Minimum(Negative Error Cost, Positive Error Cost) |
| 1 (hypothyroid) | 3 (normal) | Negative Error Cost |
| 2 (hyperthyroid) | 1 (hypothyroid) | Minimum(Negative Error Cost, Positive Error Cost) |
| 2 (hyperthyroid) | 2 (hyperthyroid) | $0.00 |
| 2 (hyperthyroid) | 3 (normal) | Negative Error Cost |
| 3 (normal) | 1 (hypothyroid) | Positive Error Cost |
| 3 (normal) | 2 (hyperthyroid) | Positive Error Cost |
| 3 (normal) | 3 (normal) | $0.00 |

## Acknowledgments

Thanks to Dr. Louise Linney for her help with interpretation of the Ontario Ministry of Health's *Schedule of Benefits*. Thanks to Martin Brooks, Grigoris Karakoulas, Cullen Schaffer, Diana Gordon, Tim Niblett, Steven Minton, and three anonymous referees of *JAIR* for their very helpful comments on earlier versions of this paper. This work was presented in informal talks at the University of Ottawa and the Naval Research Laboratory. Thanks to both audiences for their feedback.